\documentclass{article}

    \usepackage[final]{neurips_2025}

\usepackage[colorlinks=true, linkcolor=red, urlcolor=black, citecolor=blue, anchorcolor=blue, backref=page]{hyperref}
\renewcommand*{\backref}[1]{}
\renewcommand*{\backrefalt}[4]{%
    \ifcase #1%
          \or [Cited on page~#2.]%
          \else [Cited on pages~#2.]%
    \fi%
    }
\usepackage{comment}
\usepackage{natbib}
\usepackage{times}
\usepackage{latexsym}
\usepackage{bm}

\usepackage[utf8]{inputenc} % allow utf-8 input
\usepackage[T1]{fontenc}    % use 8-bit T1 fonts
\usepackage{hyperref}       % hyperlinks
\usepackage{url}            % simple URL typesetting
\usepackage{amsfonts}       % blackboard math symbols
\usepackage{nicefrac}       % compact symbols for 1/2, etc.
\usepackage{microtype}      % microtypography
\usepackage{xspace}
\usepackage{pifont}
\usepackage{tikz}
\usetikzlibrary{bayesnet}
\usepackage{wrapfig}
\usepackage{lipsum}
\usepackage{amsmath, amssymb, amsfonts}

\hypersetup{
    colorlinks=true,
    linkcolor=blue,
    }

\usepackage{array}
\usepackage{ragged2e}
\newcolumntype{C}[1]{>{\centering\arraybackslash}p{#1}}
\newcolumntype{R}[1]{>{\raggedleft\arraybackslash}p{#1}}
\newcolumntype{L}{>{\RaggedRight\arraybackslash}X}
\usepackage{makecell}

\usepackage{inconsolata}
\usepackage{amsmath}
\usepackage{mathtools}
\usepackage{graphicx}
\usepackage{centernot}
\usepackage{amsfonts}

\usepackage{enumitem}
\usepackage{soul}
\usepackage{lineno}
\usepackage{algorithm}
\usepackage{algpseudocode}
\usepackage{multirow}
\usepackage{wrapfig}
\usepackage{titletoc}

\usepackage{cleveref}
\usepackage{bbm}
\crefname{figure}{Figure}{Figures}
\crefname{table}{Table}{Tables}
\crefname{appendix}{Appendix}{Appendices}
\crefname{section}{Section}{Sections}
\crefname{equation}{Eq.}{Eqs.}
\crefname{enumi}{}{} %
\usepackage{geometry}
\usepackage{booktabs}
\usepackage{tabularx}
\usepackage{array}
\usepackage{amsmath}
\usepackage{amssymb}

\usepackage[utf8]{inputenc} % allow utf-8 input
\usepackage[T1]{fontenc}    % use 8-bit T1 fonts
\usepackage{hyperref}       % hyperlinks
\usepackage{url}            % simple URL typesetting
\usepackage{amsfonts}       % blackboard math symbols
\usepackage{nicefrac}       % compact symbols for 1/2, etc.
\usepackage{microtype}      % microtypography
\usepackage{algorithm}
\usepackage{algpseudocode}
\usepackage{amsmath}
\usepackage{ulem}
\usepackage[table]{xcolor} % Added [table] option
\definecolor{femininecolor}{rgb}{1,1,1} % Red
\definecolor{masculinecolor}{rgb}{1,1,1} % Blue
\definecolor{neutralcolor}{rgb}{1,1,1}   % Purple

\title{Subspace Mechanistic analysis}
\title{Beyond Components: Singular Vector-Based Interpretability of Transformer Circuits}

\author{
{\bf Areeb Ahmad}\thanks{Equal Contributions}
\qquad{\bf Abhinav Joshi}\footnotemark[1]\qquad {\bf Ashutosh Modi} 
 \\ 
 Indian Institute of Technology Kanpur  (IIT Kanpur) \\
  \texttt{\{ajoshi,areeb,ashutoshm\}@cse.iitk.ac.in}
}

\begin{document}

\maketitle

\begin{abstract}
Transformer-based language models exhibit complex and distributed behavior, yet their internal computations remain poorly understood. Existing mechanistic interpretability methods typically treat attention heads and multilayer perceptron layers (MLPs) (the building blocks of a transformer architecture) as indivisible units, overlooking possibilities of functional substructure learned within them. In this work, we introduce a more fine-grained perspective that decomposes these components into orthogonal singular directions, revealing superposed and independent computations within a single head or MLP. 
% This decomposition provides selective intervention, attribution, and interpretation at a level of granularity that goes beyond previous methods. 
We validate our perspective on widely used standard tasks like Indirect Object Identification (IOI), Gender Pronoun (GP), and Greater Than (GT), showing that previously identified canonical functional heads, such as the “name mover,” encode multiple overlapping subfunctions aligned with distinct singular directions. Nodes in a computational graph, that are previously identified as circuit elements show strong activation along specific low-rank directions, suggesting that meaningful computations reside in compact subspaces. While some directions remain challenging to interpret fully, our results highlight that transformer computations are more distributed, structured, and compositional than previously assumed. This perspective opens new avenues for fine-grained mechanistic interpretability and a deeper understanding of model internals.
    % Transformer-based language models exhibit complex behavior, but their internal computations remain poorly understood. Most mechanistic interpretability approaches treat components, such as attention heads and MLPs, as atomic units, ignoring potential functional substructure. We propose a finer-grained perspective that models components as superpositions of orthogonal singular directions. This perspective allows multiple independent computations to coexist within a single head or MLP, enabling selective intervention, attribution, and interpretation at a level of granularity beyond previous methods.
% We demonstrate [this approach] \AM{it looks ambiguous} on the Indirect Object Identification (IOI) \sout{task}, \Areeb{Gender Pronoun(GP) and Greater Than(GT) tasks. For instance, in IOI, we showed }\sout{showing} that well-known functional heads, like the “name mover,” encode overlapping subfunctions aligned with distinct singular directions. Nodes previously identified as part of circuits exhibit strong engagement along specific directions, supporting the view that meaningful computations are embedded in low-rank subspaces. While some functional axes remain difficult to interpret \Areeb{exhaustively}, our results reveal that transformer components are more distributed, compact, and compositional than assumed. This opens a new direction for fine-grained mechanistic interpretability and the study of model behavior.
\end{abstract}

\section{Introduction}\label{sec:intro}
% \vspace{-2mm}

Language models (LMs) exhibit complex and often surprising capabilities across tasks \citep{Radford2019LanguageMA, incontextfewshotlearners, li2023textbooksneediiphi15, javaheripi2023phi, touvron2023llama2openfoundation, jiang2023mistral7b, grattafiori2024llama3herdmodels, joshi2024cold}, yet their internal computations remain poorly understood. Mechanistic interpretability seeks to bridge this gap by identifying the circuits, i.e., networks of interacting components, that realize/show specific functions \citep{olah2020zoom, wang2022interpretabilitywildcircuitindirect}. Prior work has shown that these circuits can often be decomposed into submodules with distinct roles such as copying, inhibition, or referencing \citep{wang2022interpretabilitywildcircuitindirect}.
Despite this progress, the current state of circuit-discovery methods \citep{conmy2023towards-acdc, syed-etal-2024-attribution, bhaskar2024finding} still view model components, such as attention heads and MLPs, as atomic units of computation. Methods such as causal tracing \citep{meng2022locating}, activation patching \citep{wang2022interpretabilitywildcircuitindirect}, and attribution-based analyses \citep{heimersheim2024useinterpretactivationpatching, joshi-etal-2025-towards, joshi-etal-2025-calibration} typically probe/patch or ablate entire components to assess their functional contribution. While these techniques have produced valuable insights, they inherently assume that functionality aligns cleanly with component boundaries. In practice, however, transformer layers may multiplex multiple subfunctions within a single head or MLP, meaning that treating components as monolithic units risks overlooking the fine-grained structure of computation within these modules.

While most existing methods analyze/study transformer components as monolithic units, recent work has begun to question this assumption by investigating the internal structure of these components.
\citet{merullo2024talking}, for instance, introduced a low-rank perspective showing that attention heads communicate through specific singular directions in residual space, defined by the singular vectors of their value matrices. However, this analysis primarily captures inter-component communication, how heads “talk” to one another via low-rank channels, while leaving intra-component decomposition largely unexplored, i.e., how a single head might multiplex multiple independent functions within its internal subspace.

In this work, we extend this perspective, which goes beyond the attention heads with {additional inclusion of } MLP layers, leading to a comprehensive directional view of transformer blocks. This reveals that low-rank, distributed computations are a general feature of transformer architectures. Moreover, components identified as part of known circuits \citep{wang2022interpretabilitywildcircuitindirect, conmy2023towards-acdc, syed-etal-2024-attribution, bhaskar2024finding} exhibit strong engagement along specific singular directions, suggesting that meaningful computations are embedded within compact subspaces. We further demonstrate/validate this perspective on the widely popular canonical tasks like Indirect Object Identification (IOI) \citep{wang2022interpretabilitywildcircuitindirect}, Gender Pronoun (GP) \citep{mathwin2023identifying}, and Greater Than (GT) \citep{hanna2023does}. In IOI, for instance, our analysis identifies dominant singular directions within the same heads previously characterized as “name movers,” \citep{wang2022interpretabilitywildcircuitindirect}, showing that only a sparse subset of these directions meaningfully contributes to task performance. Using our proposed optimization scheme, we learn direction-level masks that remain highly sparse while closely replicating the model’s original behavior, indicating that transformer computations can be effectively captured by a compact set of low-rank subfunctions. Moreover, the directions corresponding to established IOI heads exhibit notably higher activation and mask weights compared to other heads, supporting the view that known circuit components operate through a small number of active, interpretable directions.
In a nutshell, we make the following contributions:
\begin{itemize}[nosep,noitemsep,leftmargin=*]
\item We introduce a directional interpretability perspective that models transformer components (attention and MLP) as superpositions of orthogonal subfunctions rather than atomic units.
\item We demonstrate this via including an optimization-based masking scheme that identifies functionally important singular directions within attention and MLP layers, enabling direction-level attribution.
\item We provide empirical evidence that multiple low-rank, interpretable computations coexist within single attention heads and MLPs, which diverge from the standard assumptions about circuit modularity.
\end{itemize}

Our findings suggest that transformer computations are not strictly modular but rather distributed, compact, and compositional, with overlapping subfunctions embedded within shared subspaces. This perspective reframes transformer interpretability through the lens of functional directions, opening a new avenue for analyzing, editing, and understanding model behavior at the subcomponent level. 

Beyond component-level decomposition, our study/investigations reveal an interesting uncovering that transformer layers naturally form stable, controllable directions in logit space, each aligned with a specific set of tokens (which we also term as \textit{\textbf{logit receptors}}, more details in Appendix \ref{app:Logit-directions}). These directions can be thought of as intrinsic mechanisms that the model selectively activates depending on the input context. For example, in gender pronoun resolution, certain directions consistently influence the logits toward tokens “\texttt{\underline{ }he}” or “\texttt{\underline{ }she},” with their activation strengths varying systematically in response to context. Importantly, we find that just scalar interventions along these directions can reliably control/modify the model’s predictions (also see Figure \ref{fig:intervention-logit-receptor}), demonstrating that these low-rank, interpretable subspaces may form interpretable building blocks of model computation. This insight provides a natural, mechanistic basis for studying how distributed computations within single heads and MLPs implement distinct functional behaviors, and motivates the fine-grained directional analysis presented in this work.

Overall, we believe that this perspective reframes how we think about transformer computations, i.e., rather than being confined to monolithic components, meaningful behaviors are often embedded in low-rank subspaces that can be independently manipulated and interpreted. This opens the door to more precise mechanistic studies, targeted model editing, and fine-grained attribution methods, and suggests that future interpretability work should consider the functional decomposition of components along intrinsic directions as a fundamental lens for understanding model behavior for abstract interpretable concepts learned by these models. We release our codebase for the experiments and additional results at \url{https://github.com/Exploration-Lab/Beyond-Components}.

% \vspace{-3mm}
\section{A Unified Linear View of Transformer Components}\label{sec:background}
% \vspace{-2mm}

To operationalize our directional interpretability perspective, we begin by expressing transformer computations in a unified linear form. 
We focus on decoder-only architectures and take the transformer circuit formulation from \citet{elhage2021mathematical} as our foundation.
By representing both attention and MLP transformations through augmented matrices that jointly include the learned weights and biases, we obtain a consistent linear representation across all components.
This “folding in” of biases (i.e., appending a constant dimension to the input and incorporating bias terms into the matrix) allows us to apply Singular Value Decomposition (SVD) uniformly to both attention and MLP layers.
{The resulting} formulation enables us to analyze orthogonal directions across different components within a shared framework, laying the groundwork for the fine-grained decomposition.

\noindent\textbf{Attention Mechanism (Query Key (QK) Interaction)} 
In a standard decoder-only transformer architecture \citep{Radford2019LanguageMA}, each attention head computes attention scores via the dot product between query and key row vectors 
\vspace{-3mm}
$$
\alpha_{ij}^{(h)} = \text{Softmax}_j\left(\frac{\mathbf{q}_i^{(h)} \cdot \mathbf{k}_j^{(h)\top}}{\sqrt{d_{\text{head}}}}\right),
$$
where \(\mathbf{q}_i^{(h)} = \mathbf{x}_i \mathbf{W}_Q^{(h)} + \mathbf{b}_Q^{(h)}\) and \(\mathbf{k}_j^{(h)} = \mathbf{x}_j \mathbf{W}_K^{(h)} + \mathbf{b}_K^{(h)}\). Expanding the dot product gives \vspace{-3mm}

$$
\mathbf{q}_i\,\mathbf{k}_j^\top = \mathbf{x}_i \mathbf{W}_Q \mathbf{W}_K^\top \mathbf{x}_j^\top + \mathbf{x}_i \mathbf{W}_Q \mathbf{b}_K^\top + \mathbf{b}_Q \mathbf{W}_K^\top \mathbf{x}_j^\top + \mathbf{b}_Q \mathbf{b}_K^\top.
$$
To express this interaction compactly as a single linear operation, we introduce an augmented matrix formulation that incorporates both weights and biases
% \vspace{-2mm}
% which can be compactly expressed using an augmented matrix
$$
[1, \mathbf{x}_i] \, \mathbf{W}_{\text{aug}}^{(QK)} \, \begin{bmatrix} 1 \\ \mathbf{x}_j^\top \end{bmatrix} = \mathbf{q}_i \cdot \mathbf{k}_j^\top.
$$
% 
% \AJ{add a line for the below}
where the augmented weight matrix is defined as
% \vspace{-2mm}
$$
\mathbf{W}_{\text{aug}}^{(QK)} =
\begin{pmatrix}
\mathbf{b}_Q \mathbf{b}_K^\top & \mathbf{b}_Q \mathbf{W}_K^\top \\
\mathbf{W}_Q \mathbf{b}_K^\top & \mathbf{W}_Q \mathbf{W}_K^\top
\end{pmatrix},
$$
% such that \vspace{-5mm}

\noindent\textbf{Attention Mechanism (Output Value (OV) Projection)} 
% After computing attention scores, the output is a weighted sum of values:
Following the computation of attention weights, each head aggregates contextual information by taking a attention weighted sum of value vectors, and projecting it into the residual stream using $\mathbf{W}_O$. This operation determines how information is written back into the residual stream:
$$
\mathbf{z}_i = \sum_j \alpha_{ij} \mathbf{v}_j, \qquad \text{where\ } \mathbf{v}_j = \mathbf{x}_j \mathbf{W}_V + \mathbf{b}_V.
$$
% each head is then projected as 
Each head’s contribution is then projected through its output matrix:
$$
\mathbf{y}_i^{(h)} = \mathbf{z}_i \mathbf{W}_O^{(h)} + \frac{1}{|H|} \mathbf{b}_O.
$$
where, \(|H|\) is the number of heads. Substituting \(\mathbf{v}_j\) and rearranging terms, the OV transformation becomes
\begin{equation}
\label{eq:ov-circuit-output}
\mathbf{y}_i^{(h)}=\sum_j \alpha_{ij} \left( \mathbf{x}_j \mathbf{W}_V \mathbf{W}_O^{(h)} + \mathbf{b}_V \mathbf{W}_O^{(h)} \right) + \frac{1}{|H|} \mathbf{b}_O
= [1 \,\, , \sum_j \alpha_{ij}\mathbf{x}_j] \,
\begin{pmatrix}
\mathbf{b}_V \mathbf{W}_O^{(h)} + \frac{1}{|H|} \mathbf{b}_O \\
\mathbf{W}_V \mathbf{W}_O^{(h)}
\end{pmatrix}
\end{equation}

We thus define the augmented output matrix for each head as
$$
\mathbf{W}_{\text{aug}}^{(OV)} = 
\begin{pmatrix}
\mathbf{b}_V \mathbf{W}_O^{(h)} + \frac{1}{|H|} \mathbf{b}_O \\
\mathbf{W}_V \mathbf{W}_O^{(h)}
\end{pmatrix}
\in \mathbb{R}^{(1 + d_{\text{model}}) \times d_{\text{model}}}.
$$

\noindent\textbf{MLP Layer Reformulation}
Beyond attention, transformer MLP blocks also consist of two affine transformations separated by a nonlinearity. To maintain a consistent linear treatment across all components, we explicitly separate these two projections and represent each using augmented matrices that include both weights and biases:
$$
\mathbf{y}_1 = \mathbf{x} \mathbf{W}_{\text{in}} + \mathbf{b}_{\text{in}}, \quad
\mathbf{y}_{\text{out}} = f(\mathbf{y}_1) \mathbf{W}_{\text{out}} + \mathbf{b}_{\text{out}}.
$$
% We further augment the weight matrices as
The augmented representations are defined as:
$$
\mathbf{W}_{\text{aug}}^{(\text{in})} =
\begin{pmatrix}
\mathbf{b}_{\text{in}} \\
\mathbf{W}_{\text{in}}
\end{pmatrix}, \quad
\mathbf{W}_{\text{aug}}^{(\text{out})} =
\begin{pmatrix}
\mathbf{b}_{\text{out}} \\
\mathbf{W}_{\text{out}}
\end{pmatrix}.
$$
% Thus, both pre- and post-activation stages of the MLP can be treated as linear maps over augmented inputs. 
Thus, both pre-activation and post-activation projections can be expressed as linear maps over augmented input vectors \([1,x]\), yielding a unified affine-to-linear transformation consistent with our treatment of attention layers.
% The above reformulations lead us to the following linear maps: 1)  $\mathbf{W}_{\rm aug}^{(QK)}\in\mathbb{R}^{(1+d_{\rm model})\times(1+d_{\rm model})}$: For attention score computation (\ref{qk_aug}); 2)
%     $\mathbf{W}_{\rm aug}^{(OV)}\in\mathbb{R}^{(1+d_{\rm model})\times(d_{\rm model})}$: For weighted value to head output(\ref{ov_aug});
%     3) $\mathbf{W}_{\rm aug}^{(in)}\in\mathbb{R}^{(1+d_{\rm model})\times d_{\rm mlp}}$ : Pre-activation MLP linear projection, and 
%     4) $\mathbf{W}_{\rm aug}^{(out)}\in\mathbb{R}^{(1+d_{\rm mlp})\times d_{\rm model}}$ : Post-activation MLP linear projection. 
% This reformulation enables us to treat all components uniformly and apply low-rank analysis, allowing us to dissect and interpret internal functionality with finer resolution.
% The above reformulations lead us to the following linear maps for various components in a transformer block:
% \medskip
% \noindent 

This unified formulation enables all major transformer subcomponents to be represented as linear operators over augmented spaces, summarized as:
% \vspace{-1mm}
\begin{align*}
    \mathbf{W}_{\text{aug}}^{(QK)} &\in \mathbb{R}^{(1 + d_{\text{model}}) \times (1 + d_{\text{model}})} 
    &&\text{: for attention score computation } \\
    \mathbf{W}_{\text{aug}}^{(OV)} &\in \mathbb{R}^{(1 + d_{\text{model}}) \times d_{\text{model}}}
    &&\text{: for weighted input-to-output transformation} \\
    \mathbf{W}_{\text{aug}}^{(\text{in})} &\in \mathbb{R}^{(1 + d_{\text{model}}) \times d_{\text{mlp}}}
    &&\text{: for MLP input projection} \\
    \mathbf{W}_{\text{aug}}^{(\text{out})} &\in \mathbb{R}^{(1 + d_{\text{mlp}}) \times d_{\text{model}}}
    &&\text{: for MLP output projection}
\end{align*}

By expressing all these transformations in a common linear framework, we can perform low-rank analyses such as SVD across both attention and MLP components in a consistent manner. This perspective lays the groundwork for the directional interpretability approach introduced in the next section, where we analyze how specific singular directions correspond to distinct, functionally meaningful computations.

% \vspace{-3mm}
\section{Directional Masking via Low-Rank Decomposition}\label{sec:methodology}
% \vspace{-2mm}

Having established a unified linear formulation of transformer components (\S\ref{sec:background}), we now turn to identifying the specific functional directions that drive model behavior.
Our goal is to decompose each augmented attention or MLP matrix into a set of orthogonal directions, each representing an independent computational axis, and to selectively intervene on these axes to understand their roles.

Rather than treating attention heads or MLP layers as monolithic units, we perform decomposition \sout{directly} on their augmented matrices, ensuring faithfulness to the model’s native computation flow while enabling fine-grained directional attribution and masking.
This formulation makes it possible to characterize a component’s behavior not in terms of entire weight matrices, but in terms of a small number of 
low-rank directions.

\noindent\textbf{Singular Value Decomposition (SVD):}
Any real matrix \( M \in \mathbb{R}^{m \times n} \) admits a singular value decomposition $ M = U\,\Sigma\,V^\top$, 
where \( U \in \mathbb{R}^{m \times m} \) and \( V \in \mathbb{R}^{n \times n} \) are orthogonal matrices, and \( \Sigma \in \mathbb{R}^{m \times n} \) is a diagonal matrix of non-negative singular values \( \sigma_1 \geq \sigma_2 \geq \cdots \geq 0 \). 
This makes singular vectors a natural coordinate system for direction-level interpretability, as they provide orthonormal bases that isolate independent computational units 
embedded within a component.

\noindent\textbf{Masking Directions:} 
To isolate and investigate the contributions of individual computational directions, we apply SVD to the augmented matrices of each attention head and MLP layer:\vspace{-5mm}
% We apply SVD to the augmented matrices of each attention head and MLP layer:

% \begin{align*}
%     \mathbf{W}_{\mathrm{aug}}^{(QK)} &= U_{QK} \, \Sigma_{QK} \, V_{QK}^\top, \\
%     \mathbf{W}_{\mathrm{aug}}^{(OV)} &= U_{OV} \, \Sigma_{OV} \, V_{OV}^\top, \\
%     \mathbf{W}_{\mathrm{aug}}^{(\text{in})} &= U_{\text{in}} \, \Sigma_{\text{in}} \, V_{\text{in}}^\top, \\
%     \mathbf{W}_{\mathrm{aug}}^{(\text{out})} &= U_{\text{out}} \, \Sigma_{\text{out}} \, V_{\text{out}}^\top.
% \end{align*}
% 
% \noindent
\begin{minipage}[t]{0.48\linewidth}
\begin{align*}
    \mathbf{W}_{\mathrm{aug}}^{(QK)} &= U_{QK} \, \Sigma_{QK} \, V_{QK}^\top, \\
    \mathbf{W}_{\mathrm{aug}}^{(\text{in})} &= U_{\text{in}} \, \Sigma_{\text{in}} \, V_{\text{in}}^\top,
\end{align*}
\end{minipage}%
\hfill
\begin{minipage}[t]{0.48\linewidth}
\begin{align*}
    \mathbf{W}_{\mathrm{aug}}^{(OV)} &= U_{OV} \, \Sigma_{OV} \, V_{OV}^\top, \\
    \mathbf{W}_{\mathrm{aug}}^{(\text{out})} &= U_{\text{out}} \, \Sigma_{\text{out}} \, V_{\text{out}}^\top,
\end{align*}
\end{minipage}

Each decomposition expresses the component as a sum of rank-1 mappings, where each term represents an orthogonal direction that can, in principle, support a distinct subfunction (also used as singular directions in the paper).
To study the importance of these subfunctions, we introduce a learnable diagonal mask matrix
\(\mathcal{M}=\mathrm{diag}(m_1,m_2, \cdots, m_r),m_i \in [0,1],\) that scales the contribution of each singular direction. The masked transformation is then defined as:
% 
% To selectively modulate contributions from individual singular directions, 
% we introduce a learnable diagonal mask matrix \AJ{add diagonal matrix notation} \( \mathcal{M} \in [0,1]^r \) 
% that scales the singular values. 
% Applying the mask yields a directionally modulated matrix:\vspace{-3mm}
\begin{align*}
    \widetilde{\mathbf{W}}_{\mathrm{aug}} 
    = U\,\Sigma\,\mathcal{M}\,V^\top,
\end{align*}
where higher entries of \( \mathcal{M} \) retain or emphasize specific directions and lower entries suppress them. 
This allows continuous, direction-level control over component behavior while maintaining differentiability for optimization.

To ensure interpretability and stability, we preserve the full representational span of the component via:
\[
U\,\Sigma\,V^\top
= U\,\Sigma\,\mathcal{M}\,V^\top 
+ U\,\Sigma\,(\mathcal{I}-\mathcal{M})\,V^\top
\]
This decomposition allows faithful interventions, i.e., the model’s downstream layers continue to receive inputs within their expected distribution (seen while training), since the overall covariance structure of activations remains intact.
Thus, both masked and complementary subspaces are retained, enabling controlled investigation without inducing distributional drift.

\begin{align*}
    \widetilde{\mathbf{W}}_{\mathrm{aug}}^{(QK)} 
        &= 
        % \left[
            U_{QK} \, \Sigma_{QK} \, \mathcal{M}_{QK} \, V_{QK}^\top \quad\quad\quad\quad\quad\quad\quad\quad\quad\quad\quad\quad\quad\quad\quad\in \mathbb{R}^{(1+d_{\text{model}})\times(1+d_{\text{model}})}\\ 
            % \quad
           % \mathbf{0}
        % \right]^\top \\
        \widetilde{\mathbf{W}}_{\mathrm{aug}}^{(OV)} 
        &= \left[
            U_{OV} \, \Sigma_{OV} \, \mathcal{M}_{OV} \, V_{OV}^\top, \quad
            U_{OV} \, \Sigma_{OV} \, (\mathcal{I} - \mathcal{M}_{OV}) \, V_{OV}^\top
        \right]^\top \quad\in \mathbb{R}^{(2(1+d_{\text{model}}))\times(d_{\text{model}})}\\
% \end{align*}
% \end{minipage}%
% \hfill
% \begin{minipage}[t]{0.48\linewidth}
% \begin{align*}
       \widetilde{\mathbf{W}}_{\mathrm{aug}}^{(\text{in})} 
        &= \left[
            U_{\text{in}} \, \Sigma_{\text{in}} \, \mathcal{M}_{\text{in}} \, V_{\text{in}}^\top, \quad
            U_{\text{in}} \, \Sigma_{\text{in}} \, (\mathcal{I} - \mathcal{M}_{\text{in}}) \, V_{\text{in}}^\top
        \right]^\top\quad\quad\quad\quad\quad\quad\in \mathbb{R}^{(2(1+d_{\text{model}}))\times(d_{\text{mlp}})} \\
    \widetilde{\mathbf{W}}_{\mathrm{aug}}^{(\text{out})} 
        &= \left[
            U_{\text{out}} \, \Sigma_{\text{out}} \, \mathcal{M}_{\text{out}} \, V_{\text{out}}^\top, \quad
            U_{\text{out}} \, \Sigma_{\text{out}} \, (\mathcal{I} - \mathcal{M}_{\text{out}}) \, V_{\text{out}}^\top
        \right]^\top\quad\quad\quad\quad\in \mathbb{R}^{(2(1+d_{\text{mlp}}))\times(d_{\text{model}})}
\end{align*}
This procedure provides a principled mechanism for identifying the causally relevant subspaces within each transformer component, while keeping the model’s computations structurally and statistically consistent with its pre-trained dynamics.

Note that for the \( \mathrm{QK} \) matrices, we retain only the masked component \( U_{QK}\,\Sigma_{QK}\,\mathcal{M}_{QK}\,V_{QK}^\top \) 
and omit the complementary term \( U_{QK}\,\Sigma_{QK}\,(\mathcal{I}-\mathcal{M}_{QK})\,V_{QK}^\top \).
This asymmetry arises from the distinct functional role of the QK block, whereas the OV and MLP matrices operate on feature representations,  the QK matrices parameterize the {attention kernel} 
, the quadratic form that defines pairwise token similarities.
Introducing a complementary \( (\mathcal{I}-\mathcal{M}) \) 
subspace here would correspond to defining additional, independent similarity maps within the same attention head. Because attention scores are normalized via a single softmax, this would implicitly produce multiple incompatible superimposed kernels,
% (and potentially up to four distinct similarity maps), 
leading to interference and spurious correlations. {In other words, if the true token representations \( x=\{x_i,x_j\} \) become correlated with irrelevant or spurious features \( x^{\mathrm{corr}}=\{x_i^{\mathrm{corr}},x_j^{\mathrm{corr}}\} \), (which is often the case due to their similar framing/syntactic-style of \( x_i^{\mathrm{corr}} \) ), the mixed terms that arise from partially masked QK components can yield misleading {similarity/attention} scores:
\(
s_{ij} \propto 
m\,x_i W_{QK} x_j^\top 
% + m(1-m)\,x_i W_{QK} x_j^{\mathrm{corr}\top}+ m(1-m)\,x_i^{\mathrm{corr}}W_{QK} x_j^{\top}
+ (1-m)\,x_i^{\mathrm{corr}} W_{QK} x_j^{\mathrm{corr}\top}.
\)
{For instance, if $x_i^{\mathrm{corr}} W_{QK} x_j^{\mathrm{corr}\top}$ term produces a similarity score close to $x_i W_{QK} x_j^{\top}$, the training objective will incorrectly suppress the mask value (\( m \rightarrow 0 \)) even if the component is genuinely useful for the task.
}
} 
Thus, we restrict QK masking to the primary \( \mathcal{M}_{QK} \), which preserves a single, coherent attention kernel while still permitting direction-level ablations over its singular basis.

\begin{algorithm}[t]
\caption{Directional Mask Optimization via Singular Value Decomposition
% \Areeb{Can we shift this to the appendix since this is a detail of the process described in the above paragraphs}
}
\label{alg:maskoptimization}
\begin{algorithmic}[1]
\Require Pretrained model \( f_\theta \), dataset \( \mathcal{D} = \{(x, y)\} \), sparsity coefficient \( \lambda \),
\State Initialize learnable diagonal masks 
\( \mathcal{M}_{QK}, \mathcal{M}_{OV}, \mathcal{M}_{\mathrm{in}}, \mathcal{M}_{\mathrm{out}} \in [0, 1]^{rank} \)
\ForAll{components (attention heads and MLP layers)}
    \State Construct augmented weight matrix \( \mathbf{W}_{\mathrm{aug}} \)
    \State Compute SVD: \( \mathbf{W}_{\mathrm{aug}} = U \, \Sigma \, V^\top \)
\EndFor
\State Freeze model parameters \( \theta \); keep masks \( \mathcal{M} \) learnable
\For{each batch \( (x, y) \in \mathcal{D} \)}
    \State Define helper function for joint representation:
    \[
        \mathbf{h}(x, x_{\mathrm{corrupt}}, \mathbf{W}) 
        \;=\; 
        \big[\,[1, x],\, [1, x_{\mathrm{corrupt}}]\,\big]\,\mathbf{W} 
        % \qquad \phi(x_i, \mathbf{W}, x_j) = [1, x_i] \widetilde{\mathbf{W}}_{\mathrm{aug}}^{QK,(l,h)} [1, x_j]^T
    \]
    \State Define component-specific reconstruction functions, for each layer $l$:
   \begin{align*}
    \mathbf{g}_{\text{attn}}^{(l)}
    % &= 
    % \sum_{h=1}^{|heads|}\mathbf{h}\!\left(
    %     \Big[\sum\nolimits_j 
    %     \mathrm{Softmax}_j\!\Big(\tfrac{x_i \widetilde{\mathbf{W}}_{\mathrm{aug}}^{QK,(l,h)} x_j^\top}{\sqrt{d_{head}}}\Big)x_j\Big],
    %     x_{\mathrm{corrupt}},
    %     \widetilde{\mathbf{W}}_{\mathrm{aug}}^{OV,(l,h)}
    % \right) \\
    % \mathbf{g}_{\text{mlp}}^{(l)}
    % \mathbf{g}_{\text{attn}}^{(l)}
    &= 
    \sum_{h=1}^{|heads|}\mathbf{h}\!\left(
        \Big[\sum\nolimits_j 
        \mathrm{Softmax}_j\!\Big(\tfrac{[1, x_i] \widetilde{\mathbf{W}}_{\mathrm{aug}}^{QK,(l,h)} [1, x_j]^\top}{\sqrt{d_{\mathrm{head}}}}\Big)x_j\Big],
        x_{\mathrm{corrupt}},
        \widetilde{\mathbf{W}}_{\mathrm{aug}}^{OV,(l,h)}
    \right) \\
    \mathbf{g}_{\text{mlp}}^{(l)}
    &= 
    \quad\mathbf{h}\!\left(
        \mathrm{GeLU}\!\big(\mathbf{h}(x,x_{\mathrm{corrupt}}^{in}, \widetilde{\mathbf{W}}_{\mathrm{aug}}^{\text{in},(l)})\big),\, 
        x_{\mathrm{corrupt}}^{out},\,
        \widetilde{\mathbf{W}}_{\mathrm{aug}}^{\text{out},(l)}
    \right)
\end{align*}
    \State Run forward pass through masked model using both contributions:
    \[
        p_{\mathcal{M}}(y \mid x) = 
        \mathrm{Softmax}\!\left(\mathrm{LayerNorm}\left[
            \text{embedding} 
            + \sum_l 
            \big(\mathbf{g}_{\text{attn}}^{(l)}
            + \mathbf{g}_{\text{mlp}}^{(l)}\big)
            \right]W_U + b_U 
        \right)
    \] 
    \State Compute objective:
    \[
        \mathcal{L}_{\mathcal{M}} =
        \mathrm{KL}\!\left[ p(y \mid x) \,\|\, p_{\mathcal{M}}(y \mid x) \right]
        + \lambda \, \|\mathrm{diag}(\mathcal{M})\|_1
    \]
    \State Update \( \mathcal{M} \) via gradient descent, and  and reconstruct weights \( \widetilde{\mathbf{W}}_{\mathrm{aug}} \)
\EndFor
\State \Return Learned masks \( \mathcal{M} \) and reconstructed weights \( \widetilde{\mathbf{W}}_{\mathrm{aug}} \)
\Statex \textbf{Note:} \( x \) (row vector) denotes the running forward pass activations while optimization, 
and \( x_{\mathrm{corrupt}} \) (row vector) denotes the appropriate corresponding corrupted or perturbed activations, that are fixed and obtained from a corrupted run.
\end{algorithmic}
\end{algorithm}

\noindent\textbf{Optimization Objective.}
To identify the singular directions most responsible for a model’s behavior on a given task, we optimize the learnable masks \( \mathcal{M} \) to balance faithfulness (preserving model behavior) and sparsity (selecting only a minimal subset of directions). 

Let \( p(y \mid x) \) denote the original model’s predictive distribution, 
and \( p_{\mathcal{M}}(y \mid x) \) the distribution obtained after applying the masked matrices. 
We define the optimization objective as:
\begin{equation*}
    \mathcal{L}_{\mathcal{M}} 
    = \mathrm{KL}\!\left[ p(y \mid x) \,\|\, p_{\mathcal{M}}(y \mid x) \right] 
      + \lambda \, \|\mathrm{diag}(\mathcal{M})\|_1,
\end{equation*}
where the KL divergence term encourages the masked model to reproduce the original model’s behavior, and the \( \ell_1 \)-regularization term promotes sparsity in the mask, selecting only a minimal subset of singular directions. 
The trade-off coefficient \( \lambda \) governs the balance between 
behavioral consistency {(faithfulness)} and sparsity. 

This procedure (summarized in Algorithm~\ref{alg:maskoptimization}) yields a low-rank, sparse decomposition, where each retained singular direction captures a distinct, functionally meaningful axis of computation within the model’s layers. While we use \( \ell_1 \)-based sparsity here, \( \ell_0 \)-regularization could provide an alternative sparse selection mechanism \citep{bhaskar2024finding,sung2021trainingneuralnetworksfixed}, and is left for future work.

\begin{figure}[t]  % figure* spans multiple columns, [p] for page float
    \centering
    \makebox[\linewidth][c]{  % centers and allows content wider than \textwidth
        \includegraphics[width=0.9\linewidth]{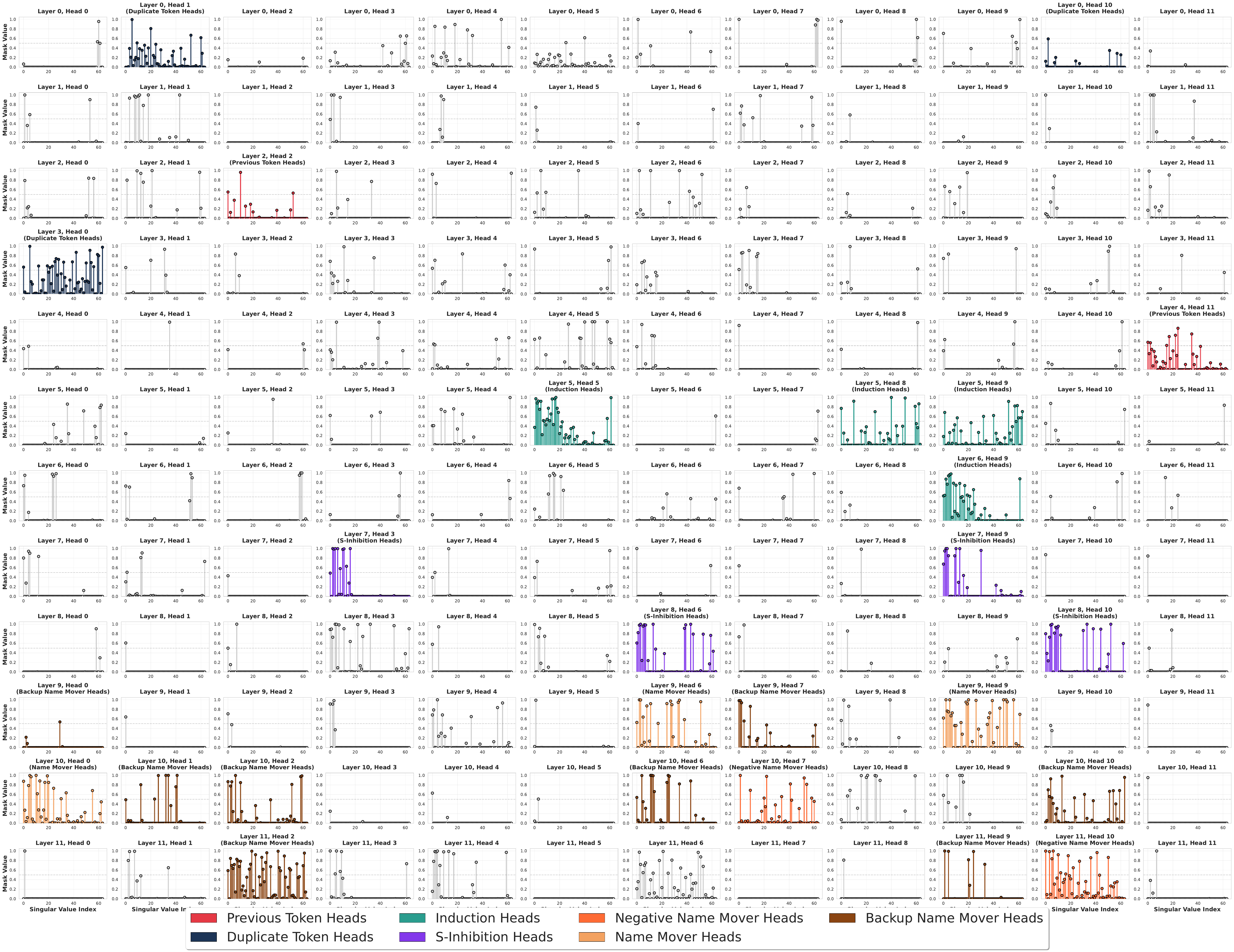}
    }
    \caption{Learned singular value masks for Query-Key ($\mathbf{W_{aug}^{QK}}$) matrices across all attention heads in the model. 
    High mask activations correspond to circuit components previously identified for the IOI task \citep{wang2022interpretabilitywildcircuitindirect}. Each head exhibits sparsity along its singular directions, revealing the fine-grained subspaces driving task behavior.
    }
    \label{fig:qk_singular_masks}
    \vspace{-7mm}
\end{figure}

To improve robustness and better isolate task-relevant directions, we feed the model concatenated activations from both clean and corrupted inputs, where \( x \) denotes the clean activations from the original input, 
and \( x_{\mathrm{corrupt}} \) denotes corresponding corrupted or perturbed activations (e.g., with noise or prompt that should consist of
a small change to x that would result in a different label in the task). 
The combination \( x,\, x_{\mathrm{corrupt}} \) forms a joint input, leading to the next token prediction {\( p_{\mathcal{M}}(y \mid x) \) $=f\big(\sum_{k} (m_kx+(1-m_k)x_{corrupt})\sigma_ku_kv_k^\top\big)$ (details in Algorithm \ref{alg:maskoptimization})}, enabling the mask optimization to identify singular directions that robustly capture task-relevant computations.

% \vspace{-3mm}
\section{Experiments}\label{sec:experiments}
% \vspace{-2mm}

Having introduced directional masking, which decomposes each transformer component into orthogonal singular directions and enables selective interventions, we now ask: \textit{do these fine-grained directions correspond to meaningful, task-relevant subfunctions within the model?}

Prior interpretability studies have associated specific attention heads or MLP layers with distinct behaviors, such as syntactic role tracking or inhibition \citep{wang2022interpretabilitywildcircuitindirect, conmy2023towards-acdc}. However, these analyses treat components as atomic units. In contrast, our approach allows us to examine computation at the level of individual singular directions, revealing whether components multiplex multiple independent computations along distinct low-rank axes.

We apply our method to a pretrained GPT-2 Small model \citep{Radford2019LanguageMA}, a tractable benchmark widely used in mechanistic interpretability \citep{wang2022interpretabilitywildcircuitindirect, hanna2023does}. To inspect the generality of directional subfunctions, we evaluate the model on three representative tasks: Indirect Object Identification (IOI) \citep{wang2022interpretabilitywildcircuitindirect}, which tests syntactic reasoning by examining coreference resolution in sentences. 
% \sout{like \textit{“Friends Juana and Kristi found a mango at the bar. Kristi gave it to $\rightarrow$ Juana”}, where identifying the indirect object requires grammatical understanding}. 
Greater-Than (GT) \citep{hanna2023does}, where the model predicts numerical tokens following a number in context, assessing its quantitative reasoning. Gender Pronoun Resolution (GP) \citep{mathwin2023identifying}, which measures semantic reasoning by testing pronoun-to-antecedent resolution in natural text. Full dataset details are provided in App.~\ref{app:sec-real-dataset}.

We organize our experiments around four core research questions. \textbf{R1) Can} a small number of learned singular directions faithfully preserve model behavior? \textbf{R2) Do} these directions align with components identified in prior circuit-level analyses? \textbf{R3) Can} our method decompose known functional heads into interpretable sub-functions? \textbf{R4) Can} we discover new functional axes not identifiable through standard component-level analysis?

For each attention and MLP component in the GPT-2 Small model, we extract its augmented matrices {$\mathbf{W}_{\mathrm{aug}}^{(QK)},\mathbf{W}_{\mathrm{aug}}^{(OV)},\mathbf{W}_{\mathrm{aug}}^{(in)}\text{ and, }\mathbf{W}_{\mathrm{aug}}^{(out)}$} (as described in \S\ref{sec:background}) and perform a singular value decomposition.
We retain all non-zero singular directions, corresponding to the actual matrix rank rather than a fixed truncation. For the Query--Key (\(\mathbf{W}_{\mathrm{aug}}^{QK}\)) matrices, the effective rank is 64, hence \(r = 64\) reflects the full set of functional directions. Similarly, the Output-Value (\(\mathbf{W}_{\mathrm{aug}}^{OV}\)) matrices exhibit a rank of 65 after augmentation, while MLP layers typically retain their full non-zero spectrum. Empirically, we found these truncations yield minuscule drops (in the range of ($1e-6$) in reconstruction faithfulness (KLD), confirming that the effective ranks of the augmented matrices capture all functionally relevant subspaces.

Over these singular directions, we introduce a learnable diagonal mask that scales the singular values, enabling direction-level modulation of each component’s contribution. The masks are optimized using the objective defined in \S\ref{sec:methodology}.
% \sout{with an additional $\ell_1$ regularization term to promote sparsity and isolate the most behaviorally relevant directions}. 
The optimization in Algorithm~\ref{alg:maskoptimization} proceeds on mini-batches drawn from each dataset, and optimization is stopped early once the held-out reconstruction loss stabilizes. For corruption (\(x_{\mathrm{corrupt}}\)), we follow the datasets provided by \citet{bhaskar2024finding}, and create an additional set of datapoints for tasks GT and GP.

\begin{table}[t]
\centering
\caption{Comparison of sparsity, reconstruction fidelity, and task performance across datasets. Sparsity is measured relative to the number of non-zero singular directions in each component (“Relative”) and as a fraction of the full matrix size (“Full”). KL divergence quantifies reconstruction loss, while accuracy (where applicable) and exact match show downstream task performance. Despite extreme sparsity, the learned directions retain high behavioral fidelity. (see App. \ref{app-sec:sparsity} for sparsity computations).
% \Areeb{Sparsity approximate since we don't do hard cutting off}
} 
\begin{tabular}{ccccc}
\toprule
\textbf{Dataset} & \textbf{Sparsity (Rel / Full)} & \textbf{KLD} & \textbf{Acc. (Pruned / Full)} & \textbf{Exact Match} \\
\midrule
IOI & 91.32 / 98.66 & 0.21$\pm$0.02 & 0.70$\pm$0.07 / 0.79$\pm$0.05 & 0.77$\pm$0.06 \\
GT  & 95.21 / 99.26 & 0.23$\pm$0.03 & N/A &  0.33$\pm$0.06 \\
GP  & 96.81 / 99.51 & 0.13$\pm$0.01 & 0.75$\pm$0.04 / 0.77$\pm$0.04 & 0.86$\pm$0.07 \\
\bottomrule
\end{tabular}
\label{tab:sparsity_results}
% \vspace{-5mm}
\end{table}

Our first experiment evaluates whether the model’s behavior can be faithfully reconstructed using only a small subset of learned singular directions (Table~\ref{tab:sparsity_results}).
Remarkably, across all datasets, high-fidelity reconstruction is possible with far fewer directions than the full component. 
For instance, the IOI task retains only \(\sim 9\%\) of directions relative to the full component while achieving a KL divergence of 0.21 and an exact match of 0.77, demonstrating that a small fraction of singular directions suffices to reproduce the model’s behavior. 
Similar patterns hold for GT and GP, where over 95\% of directions can be pruned with minimal impact on performance.
The learned masks are distinctly sparse, activating only a handful of directions per layer. \begin{wrapfigure}{r}{0.5\textwidth}
% \begin{figure}[t]
\centering
 \includegraphics[width=0.48\textwidth]{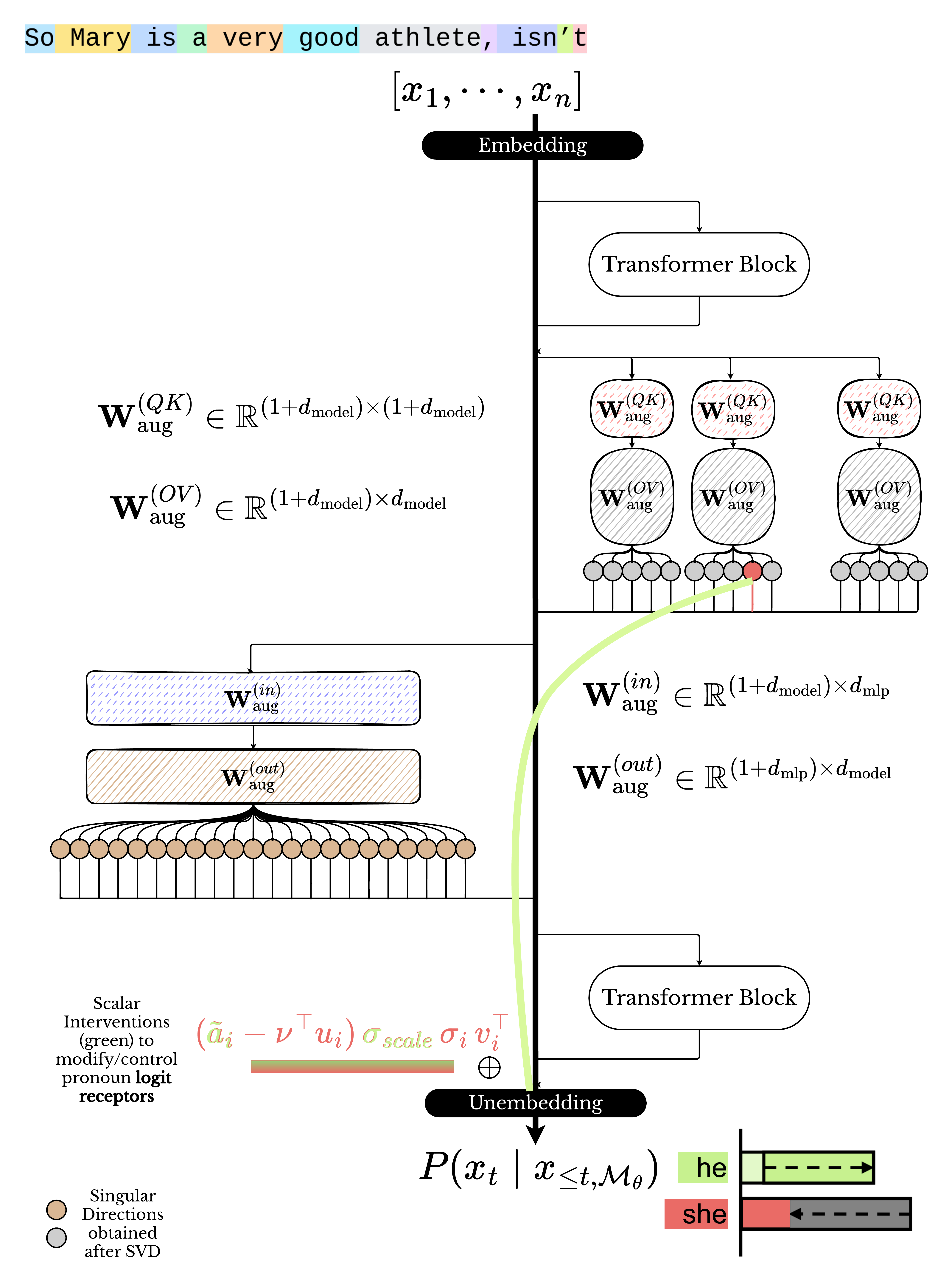}
  \caption{
  The Figure shows intervention in the logit receptor for the Gender Pronoun task. Controlling the logit receptor using a scalar intervention modifies the predicted logits. }
  \label{fig:intervention-logit-receptor}
  \vspace{-5mm}
% \end{figure}
\end{wrapfigure}
These selected directions achieve substantially lower KL divergence than top-k magnitude or random SVD baselines, suggesting that task-relevant computation is concentrated along specific, semantically meaningful directions rather than the largest singular modes (quantitative details in App.~\ref{app:additional-results-section}).

Next, we investigate how our learned singular directions correspond to previously identified mechanistic circuits, using the IOI task as a reference and comparing to ACDC \citep{conmy2023towards-acdc} and \citet{wang2022interpretabilitywildcircuitindirect}. While prior analyses treat entire attention heads as circuit participants, our direction-level decomposition reveals that most heads exhibit strong activation along only a few singular directions. 
In other words, the coarse, component-level circuits reported in previous work arise from finer, low-rank structures embedded within each head.

Figures~\ref{fig:qk_singular_masks}, \ref{fig:ov_singular_masks}, and \ref{fig:mlp_singular_masks} visualize the learned singular-value masks for Query-Key, OV, and MLP matrices, respectively, across the full model. Heads known to play key roles in IOI, such as Name Mover, Backup Name Mover, and S-Inhibition \cite{wang2022interpretabilitywildcircuitindirect}, show consistently high activations across multiple singular directions, pinpointing the precise subspaces driving their behavior. Conversely, components not associated with known circuits exhibit near-zero activations across all directions (App. Figure~\ref{fig:attention-mask-distribution}), demonstrating that our optimization selectively isolates task-relevant subspaces.

Overall, these results demonstrate that transformer components are not monolithic, and their internal computation is distributed along a small number of interpretable, low-rank axes/bases. The presented perspective thus bridges component-level circuit discovery and fine-grained mechanistic analysis, providing a new lens to study how distinct computational roles coexist within the same architectural unit.

Extending this perspective, we find that transformer layers {contain inherent} fixed directions in logit space corresponding to stable token-preference axes. 
Task behavior emerges from dynamically steering these directions via input-dependent scalar activations; in the Gender Pronoun task, for instance, distinct “he” and “she” directions exist, and the model selectively activates them depending on context. Scalar-based interventions confirm that these directions are causally relevant by flipping gender pronoun predictions with perfect accuracy, demonstrating that low-rank subspaces serve as modular, functional building blocks bridging representation and output behavior. All these results provide a unified mechanistic view, i.e., computation is concentrated along sparse, interpretable singular directions within components, which in turn map to stable, task-relevant axes in logit space (full details in App. \ref{app:Logit-directions}).

% % \vspace{-3mm}
\section{Functional Decomposition of Attention Head 9.6 via Singular Directions}
\label{sec:head_96_analysis}
% % \vspace{-2mm}

Building on our discovery that transformer components distribute computation along low-rank, interpretable directions, we next ask: what specific operations does a single head implement through these directions? We focus on Head 9.6, previously identified as a key contributor to the IOI task \citep{wang2022interpretabilitywildcircuitindirect, conmy2023towards-acdc, bhaskar2024finding}, and examine its singular directions with high learned mask values.
Our analysis reveals 
% \sout{that this head does not perform a monolithic computation. Instead, }
individual singular directions encode distinct, separable operations, ranging from syntactic cues to semantic entity tracking, that collectively support task performance. Table~\ref{tab:functional_role} summarizes the primary roles and their corresponding mask values, highlighting how these directions act as interpretable computational primitives within a single attention head.

\noindent\textbf{Semantic Separation of Entities and Actions}
Some singular directions carve out conceptually meaningful subspaces, revealing that attention heads implement abstract computations rather than just capturing statistical variance. A striking example is the $7^{\text{th}}$ singular direction ($S_7$) in Head 9.6, which consistently separates named entities (e.g., “Mary”, “Kevin”) from action-related tokens (e.g., “went”, “gave”). As shown in Table~\ref{tab:entity_action_seperation}, entity tokens exhibit strong positive activations {($+3.52 \pm 1.42$)}, while action tokens are suppressed {($-4.44 \pm 0.68$)}. For instance, in the sentence “Jerry and Mary went to the school. Mary gave a raspberry to”, $S_7$ assigns high activations to “Jerry” ($+2.87$) and “Mary” ($+2.78$), but negative values to “went” ($-4.33$) and “gave” ($-4.17$).
This direction acts like a semantic filter, splitting “who” from “what is being done,” creating a foundation upon which downstream heads can operate.

\noindent\textbf{Entity Salience and Detection}
Another key computation is performed by the $28^{\text{th}}$ singular direction ($S_{28}$), which detects and amplifies the salience of named entities. In IOI prompts, tracking participants across multiple mentions is critical, and $S_{28}$ highlights entities regardless of their position or specific lexical form. Table~\ref{tab:entity_detection} shows that named entities receive high activations (e.g., “Susan”: $4.05$, “Kevin”: $5.22$), whereas function words like “the” and “of” remain low. Interestingly, first mentions often get stronger activations than subsequent mentions (e.g., “Kevin”: $5.22$, “Kevin$_2$: $2.07$), suggesting this direction is sensitive to positional salience.
We interpret $S_{28}$ as an entity salience signal, priming tokens for further grammatical and referential reasoning.

\noindent\textbf{Sequence Initialization Detection}
The top singular direction ($S_1$) implements a structural, positional primitive, i.e., it assigns extremely high positive values to the first token in a sequence while giving negligible or negative values to all others. Across prompts, the first token receives activations 20–25\(\times\) larger than subsequent tokens.
This behavior is not unique to Head 9.6; similar patterns appear in other heads and layers, independent of task, e.g., {$S_1$ of} Head 0 in Layer 10, suggesting that sequence initialization is a reusable primitive distributed across the network.

\noindent\textbf{Summary and Implications}
All these examples show that Head 9.6 multiplexes multiple independent computations through orthogonal singular directions: semantic discrimination ($S_7$), entity salience ($S_{28}$), and sequence initialization ($S_1$). This supports the hypothesis that transformer components are integration points of overlapping, low-rank subfunctions, rather than monolithic units.
The directional perspective provides a fine-grained lens for mechanistic interpretability, isolating the entangled computations within attention heads that would otherwise appear inseparable. Figure~\ref{fig:function-svd-9.6} summarizes the functional roles identified. Additional findings in App.~\ref{app:additional-results-section} reinforce the idea that circuits should be understood not at the level of whole heads, but as a collection of directions, each contributing in distinct proportions to task-relevant behavior.

\section{Related Work}
\label{sec:relatedwork}

Mechanistic interpretability seeks to break down neural networks into human-understandable components. Prior work has uncovered specialized attention heads and MLP circuits supporting tasks like indirect object identification and fact recall, typically treating heads or layers as atomic units \citep{elhage2021mathematical, wang2022interpretabilitywildcircuitindirect, meng2022locating}.
Our work complements this by going inside each head using singular value decomposition (SVD) on augmented weight matrices. This reveals independent computational directions, each implementing a distinct function, such as semantic separation, entity salience, or sequence initialization in head 9.6. While past SVD-based analyses focused on isolated components \citep{gao2024scalingevaluatingsparseautoencoders, cunningham2023sparseautoencodershighlyinterpretable}, we generalize the approach across all core circuits, query-key, output-value, and MLP transformations, showing that individual directions, rather than entire heads, can carry task-relevant computations.
This perspective also complements studies showing interdependencies across heads \citep{merullo2024circuitcomponentreusetasks}. Instead of modeling communication between heads, we decompose each head internally, highlighting low-rank primitives that drive behavior. In doing so, we introduce a direction-level granularity, which provides a finer, mechanistically meaningful lens on model computation and opens a path toward understanding how overlapping subfunctions combine within a single attention head.
% \vspace{-3mm}
\section{Discussion and Limitations}
\label{sec:limitations}
% \vspace{-2mm}

\noindent\textbf{Rethinking Transformer Components.}
Our work emphasizes a simple but powerful idea: transformer components are not monolithic units, but rather collections of independent functional directions. Each direction can encode a distinct computational primitive, such as sequence initialization, entity salience, or semantic separation, allowing a single head or MLP block to multiplex multiple sub-functions. By zooming in on these low-rank directions, we can reveal hidden structure that is invisible when treating heads or layers as atomic, providing a finer-grained understanding of how transformers compute.
This perspective has broad implications. It suggests that mechanistic interpretability should move beyond unit-level analysis, toward frameworks that capture both the individual roles of directions and their interactions, potentially defining new “micro-circuits” within components. Future work could explore direction-level communication patterns, drawing inspiration from recent studies on head-to-head interaction graphs \citep{merullo2024talking}, to understand how these sub-functions combine into higher-level behaviors.

\noindent\textbf{Limitations:}
The directional decomposition perspective we present provides a fine-grained view into transformer computations; however, several limitations remain for further exploration. First, our analysis considers each augmented matrix independently, focusing on individual layers and heads. This isolation facilitates precise attribution of function to specific singular directions but may obscure emergent behaviors that arise from interactions across components, which are principal to many transformer computations. Second, the approach assumes that singular directions correspond to interpretable and causally relevant subroutines. Although supported by observed semantic boundaries and activation patterns, this assumption lacks formal justification, and task-relevant computation may be distributed across weaker directions or higher-order interactions. Third, our method applies learnable diagonal masks to fixed singular vectors, thereby restricting optimization to axis-aligned subspaces, which may limit expressiveness. Allowing controlled perturbations of singular directions could capture finer task-dependent variations. Fourth, our evaluation is limited to the standard benchmark tasks and GPT-2 Small. Whether the method scales to more complex reasoning tasks, larger models, or instruction-tuned systems remains an open question. 
Finally, while our method reveals functionally interpretable directions, it does not yet fully elucidate the causal role of each direction in the end-to-end model behavior. Our masking objective encourages output faithfulness, but disentangling true causal mediation from representational correlation remains an open challenge. Future work could integrate causal tracing or activation patching to more directly establish mechanistic influence.

Despite these limitations, directional decomposition provides a new lens on transformer internals. By treating model components as collections of functional directions, we can uncover the computational primitives that underlie complex behaviors, laying the foundation for more systematic and fine-grained circuit discovery in modern language models.

\vspace{-3mm}
\section{Conclusion}
\label{sec:conclusion}
\vspace{-2mm}
Transformers are often seen as black boxes, complex computational graphs of neurons, heads, and layers whose inner workings are difficult to disentangle. In this work, we demonstrate that a simple shift in perspective, focusing on individual singular directions in parameter space, can reveal surprisingly clean and interpretable sub-computations.
This perspective also suggests a broader narrative. 
Might we someday construct compact, modular explanations of entire model behaviors by stitching together just a few interpretable directions?
If so, we may one day piece together modular explanations of entire model behaviors from
a small number of interpretable directions.

\section*{Acknowledgments}

We would like to thank the anonymous reviewers and the meta-reviewer for their insightful comments and suggestions. This research work was partially supported by the Research-I Foundation of the Department of CSE at IIT Kanpur.

\bibliography{references}

@inproceedings{
conmy2023towards-acdc,
title={Towards Automated Circuit Discovery for Mechanistic Interpretability},
author={Arthur Conmy and Augustine N. Mavor-Parker and Aengus Lynch and Stefan Heimersheim and Adri{\`a} Garriga-Alonso},
booktitle={Thirty-seventh Conference on Neural Information Processing Systems},
year={2023},
url={https://openreview.net/forum?id=89ia77nZ8u}
}

@inproceedings{
bhaskar2024finding,
title={Finding Transformer Circuits With Edge Pruning},
author={Adithya Bhaskar and Alexander Wettig and Dan Friedman and Danqi Chen},
booktitle={The Thirty-eighth Annual Conference on Neural Information Processing Systems},
year={2024},
url={https://openreview.net/forum?id=8oSY3rA9jY}
}

@article{olah2020zoom,
  author = {Olah, Chris and Cammarata, Nick and Schubert, Ludwig and Goh, Gabriel and Petrov, Michael and Carter, Shan},
  title = {Zoom In: An Introduction to Circuits},
  journal = {Distill},
  year = {2020},
  note = {https://distill.pub/2020/circuits/zoom-in},
  doi = {10.23915/distill.00024.001}
}

@article{elhage2021mathematical,
   title={A Mathematical Framework for Transformer Circuits},
   author={Elhage, Nelson and Nanda, Neel and Olsson, Catherine and Henighan, Tom and Joseph, Nicholas and Mann, Ben and Askell, Amanda and Bai, Yuntao and Chen, Anna and Conerly, Tom and DasSarma, Nova and Drain, Dawn and Ganguli, Deep and Hatfield-Dodds, Zac and Hernandez, Danny and Jones, Andy and Kernion, Jackson and Lovitt, Liane and Ndousse, Kamal and Amodei, Dario and Brown, Tom and Clark, Jack and Kaplan, Jared and McCandlish, Sam and Olah, Chris},
   year={2021},
   journal={Transformer Circuits Thread},
   note={https://transformer-circuits.pub/2021/framework/index.html}
}

@misc{wang2022interpretabilitywildcircuitindirect,
      title={Interpretability in the Wild: a Circuit for Indirect Object Identification in GPT-2 small}, 
      author={Kevin Wang and Alexandre Variengien and Arthur Conmy and Buck Shlegeris and Jacob Steinhardt},
      year={2022},
      eprint={2211.00593},
      archivePrefix={arXiv},
      primaryClass={cs.LG},
      url={https://arxiv.org/abs/2211.00593}, 
}

@misc{merullo2024circuitcomponentreusetasks,
      title={Circuit Component Reuse Across Tasks in Transformer Language Models}, 
      author={Jack Merullo and Carsten Eickhoff and Ellie Pavlick},
      year={2024},
      eprint={2310.08744},
      archivePrefix={arXiv},
      primaryClass={cs.CL},
      url={https://arxiv.org/abs/2310.08744}, 
}

@misc{heimersheim2024useinterpretactivationpatching,
      title={How to use and interpret activation patching}, 
      author={Stefan Heimersheim and Neel Nanda},
      year={2024},
      eprint={2404.15255},
      archivePrefix={arXiv},
      primaryClass={cs.LG},
      url={https://arxiv.org/abs/2404.15255}, 
}

@misc{Radford2019LanguageMA,
  title={Language Models are Unsupervised Multitask Learners},
  author={Alec Radford and Jeff Wu and Rewon Child and David Luan and Dario Amodei and Ilya Sutskever},
  year={2019},
  url={https://api.semanticscholar.org/CorpusID:160025533}
}

@article{meng2022locating,
  title={Locating and Editing Factual Associations in {GPT}},
  author={Kevin Meng and David Bau and Alex Andonian and Yonatan Belinkov},
  journal={Advances in Neural Information Processing Systems},
  volume={35},
  year={2022}
}

@inproceedings{syed-etal-2024-attribution,
    title = "Attribution Patching Outperforms Automated Circuit Discovery",
    author = "Syed, Aaquib  and
      Rager, Can  and
      Conmy, Arthur",
    editor = "Belinkov, Yonatan  and
      Kim, Najoung  and
      Jumelet, Jaap  and
      Mohebbi, Hosein  and
      Mueller, Aaron  and
      Chen, Hanjie",
    booktitle = "Proceedings of the 7th BlackboxNLP Workshop: Analyzing and Interpreting Neural Networks for NLP",
    month = nov,
    year = "2024",
    address = "Miami, Florida, US",
    publisher = "Association for Computational Linguistics",
    url = "https://aclanthology.org/2024.blackboxnlp-1.25/",
    doi = "10.18653/v1/2024.blackboxnlp-1.25",
    pages = "407--416"
}

@article{incontextfewshotlearners,
  author       = {Tom B. Brown and
                  Benjamin Mann and
                  Nick Ryder and
                  Melanie Subbiah and
                  Jared Kaplan and
                  Prafulla Dhariwal and
                  Arvind Neelakantan and
                  Pranav Shyam and
                  Girish Sastry and
                  Amanda Askell and
                  Sandhini Agarwal and
                  Ariel Herbert{-}Voss and
                  Gretchen Krueger and
                  Tom Henighan and
                  Rewon Child and
                  Aditya Ramesh and
                  Daniel M. Ziegler and
                  Jeffrey Wu and
                  Clemens Winter and
                  Christopher Hesse and
                  Mark Chen and
                  Eric Sigler and
                  Mateusz Litwin and
                  Scott Gray and
                  Benjamin Chess and
                  Jack Clark and
                  Christopher Berner and
                  Sam McCandlish and
                  Alec Radford and
                  Ilya Sutskever and
                  Dario Amodei},
  title        = {Language Models are Few-Shot Learners},
  journal      = {CoRR},
  volume       = {abs/2005.14165},
  year         = {2020},
  url          = {https://arxiv.org/abs/2005.14165},
  eprinttype    = {arXiv},
  eprint       = {2005.14165},
  bibsource    = {dblp computer science bibliography, https://dblp.org}
}

@inproceedings{
merullo2024talking,
title={Talking Heads: Understanding Inter-Layer Communication in Transformer Language Models},
author={Jack Merullo and Carsten Eickhoff and Ellie Pavlick},
booktitle={The Thirty-eighth Annual Conference on Neural Information Processing Systems},
year={2024},
url={https://openreview.net/forum?id=LUsx0chTsL}
}

@misc{cunningham2023sparseautoencodershighlyinterpretable,
      title={Sparse Autoencoders Find Highly Interpretable Features in Language Models}, 
      author={Hoagy Cunningham and Aidan Ewart and Logan Riggs and Robert Huben and Lee Sharkey},
      year={2023},
      eprint={2309.08600},
      archivePrefix={arXiv},
      primaryClass={cs.LG},
      url={https://arxiv.org/abs/2309.08600}, 
}

@misc{gao2024scalingevaluatingsparseautoencoders,
      title={Scaling and evaluating sparse autoencoders}, 
      author={Leo Gao and Tom Dupré la Tour and Henk Tillman and Gabriel Goh and Rajan Troll and Alec Radford and Ilya Sutskever and Jan Leike and Jeffrey Wu},
      year={2024},
      eprint={2406.04093},
      archivePrefix={arXiv},
      primaryClass={cs.LG},
      url={https://arxiv.org/abs/2406.04093}, 
}

@misc{jiang2023mistral7b,
      title={Mistral 7B}, 
      author={Albert Q. Jiang and Alexandre Sablayrolles and Arthur Mensch and Chris Bamford and Devendra Singh Chaplot and Diego de las Casas and Florian Bressand and Gianna Lengyel and Guillaume Lample and Lucile Saulnier and Lélio Renard Lavaud and Marie-Anne Lachaux and Pierre Stock and Teven Le Scao and Thibaut Lavril and Thomas Wang and Timothée Lacroix and William El Sayed},
      year={2023},
      eprint={2310.06825},
      archivePrefix={arXiv},
      primaryClass={cs.CL},
      url={https://arxiv.org/abs/2310.06825}, 
}

@misc{touvron2023llama2openfoundation,
      title={Llama 2: Open Foundation and Fine-Tuned Chat Models}, 
      author={Hugo Touvron and Louis Martin and Kevin Stone and Peter Albert and Amjad Almahairi and Yasmine Babaei and Nikolay Bashlykov and Soumya Batra and Prajjwal Bhargava and Shruti Bhosale and Dan Bikel and Lukas Blecher and Cristian Canton Ferrer and Moya Chen and Guillem Cucurull and David Esiobu and Jude Fernandes and Jeremy Fu and Wenyin Fu and Brian Fuller and Cynthia Gao and Vedanuj Goswami and Naman Goyal and Anthony Hartshorn and Saghar Hosseini and Rui Hou and Hakan Inan and Marcin Kardas and Viktor Kerkez and Madian Khabsa and Isabel Kloumann and Artem Korenev and Punit Singh Koura and Marie-Anne Lachaux and Thibaut Lavril and Jenya Lee and Diana Liskovich and Yinghai Lu and Yuning Mao and Xavier Martinet and Todor Mihaylov and Pushkar Mishra and Igor Molybog and Yixin Nie and Andrew Poulton and Jeremy Reizenstein and Rashi Rungta and Kalyan Saladi and Alan Schelten and Ruan Silva and Eric Michael Smith and Ranjan Subramanian and Xiaoqing Ellen Tan and Binh Tang and Ross Taylor and Adina Williams and Jian Xiang Kuan and Puxin Xu and Zheng Yan and Iliyan Zarov and Yuchen Zhang and Angela Fan and Melanie Kambadur and Sharan Narang and Aurelien Rodriguez and Robert Stojnic and Sergey Edunov and Thomas Scialom},
      year={2023},
      eprint={2307.09288},
      archivePrefix={arXiv},
      primaryClass={cs.CL},
      url={https://arxiv.org/abs/2307.09288}, 
}

@misc{grattafiori2024llama3herdmodels,
      title={The Llama 3 Herd of Models}, 
      author={Aaron Grattafiori and Abhimanyu Dubey and Abhinav Jauhri and Abhinav Pandey and Abhishek Kadian and Ahmad Al-Dahle and Aiesha Letman and Akhil Mathur and Alan Schelten and Alex Vaughan and Amy Yang and Angela Fan and Anirudh Goyal and Anthony Hartshorn and Aobo Yang and Archi Mitra and Archie Sravankumar and Artem Korenev and Arthur Hinsvark and Arun Rao and Aston Zhang and Aurelien Rodriguez and Austen Gregerson and Ava Spataru and Baptiste Roziere and Bethany Biron and Binh Tang and Bobbie Chern and Charlotte Caucheteux and Chaya Nayak and Chloe Bi and Chris Marra and Chris McConnell and Christian Keller and Christophe Touret and Chunyang Wu and Corinne Wong and Cristian Canton Ferrer and Cyrus Nikolaidis and Damien Allonsius and Daniel Song and Danielle Pintz and Danny Livshits and Danny Wyatt and David Esiobu and Dhruv Choudhary and Dhruv Mahajan and Diego Garcia-Olano and Diego Perino and Dieuwke Hupkes and Egor Lakomkin and Ehab AlBadawy and Elina Lobanova and Emily Dinan and Eric Michael Smith and Filip Radenovic and Francisco Guzmán and Frank Zhang and Gabriel Synnaeve and Gabrielle Lee and Georgia Lewis Anderson and Govind Thattai and Graeme Nail and Gregoire Mialon and Guan Pang and Guillem Cucurell and Hailey Nguyen and Hannah Korevaar and Hu Xu and Hugo Touvron and Iliyan Zarov and Imanol Arrieta Ibarra and Isabel Kloumann and Ishan Misra and Ivan Evtimov and Jack Zhang and Jade Copet and Jaewon Lee and Jan Geffert and Jana Vranes and Jason Park and Jay Mahadeokar and Jeet Shah and Jelmer van der Linde and Jennifer Billock and Jenny Hong and Jenya Lee and Jeremy Fu and Jianfeng Chi and Jianyu Huang and Jiawen Liu and Jie Wang and Jiecao Yu and Joanna Bitton and Joe Spisak and Jongsoo Park and Joseph Rocca and Joshua Johnstun and Joshua Saxe and Junteng Jia and Kalyan Vasuden Alwala and Karthik Prasad and Kartikeya Upasani and Kate Plawiak and Ke Li and Kenneth Heafield and Kevin Stone and Khalid El-Arini and Krithika Iyer and Kshitiz Malik and Kuenley Chiu and Kunal Bhalla and Kushal Lakhotia and Lauren Rantala-Yeary and Laurens van der Maaten and Lawrence Chen and Liang Tan and Liz Jenkins and Louis Martin and Lovish Madaan and Lubo Malo and Lukas Blecher and Lukas Landzaat and Luke de Oliveira and Madeline Muzzi and Mahesh Pasupuleti and Mannat Singh and Manohar Paluri and Marcin Kardas and Maria Tsimpoukelli and Mathew Oldham and Mathieu Rita and Maya Pavlova and Melanie Kambadur and Mike Lewis and Min Si and Mitesh Kumar Singh and Mona Hassan and Naman Goyal and Narjes Torabi and Nikolay Bashlykov and Nikolay Bogoychev and Niladri Chatterji and Ning Zhang and Olivier Duchenne and Onur Çelebi and Patrick Alrassy and Pengchuan Zhang and Pengwei Li and Petar Vasic and Peter Weng and Prajjwal Bhargava and Pratik Dubal and Praveen Krishnan and Punit Singh Koura and Puxin Xu and Qing He and Qingxiao Dong and Ragavan Srinivasan and Raj Ganapathy and Ramon Calderer and Ricardo Silveira Cabral and Robert Stojnic and Roberta Raileanu and Rohan Maheswari and Rohit Girdhar and Rohit Patel and Romain Sauvestre and Ronnie Polidoro and Roshan Sumbaly and Ross Taylor and Ruan Silva and Rui Hou and Rui Wang and Saghar Hosseini and Sahana Chennabasappa and Sanjay Singh and Sean Bell and Seohyun Sonia Kim and Sergey Edunov and Shaoliang Nie and Sharan Narang and Sharath Raparthy and Sheng Shen and Shengye Wan and Shruti Bhosale and Shun Zhang and Simon Vandenhende and Soumya Batra and Spencer Whitman and Sten Sootla and Stephane Collot and Suchin Gururangan and Sydney Borodinsky and Tamar Herman and Tara Fowler and Tarek Sheasha and Thomas Georgiou and Thomas Scialom and Tobias Speckbacher and Todor Mihaylov and Tong Xiao and Ujjwal Karn and Vedanuj Goswami and Vibhor Gupta and Vignesh Ramanathan and Viktor Kerkez and Vincent Gonguet and Virginie Do and Vish Vogeti and Vítor Albiero and Vladan Petrovic and Weiwei Chu and Wenhan Xiong and Wenyin Fu and Whitney Meers and Xavier Martinet and Xiaodong Wang and Xiaofang Wang and Xiaoqing Ellen Tan and Xide Xia and Xinfeng Xie and Xuchao Jia and Xuewei Wang and Yaelle Goldschlag and Yashesh Gaur and Yasmine Babaei and Yi Wen and Yiwen Song and Yuchen Zhang and Yue Li and Yuning Mao and Zacharie Delpierre Coudert and Zheng Yan and Zhengxing Chen and Zoe Papakipos and Aaditya Singh and Aayushi Srivastava and Abha Jain and Adam Kelsey and Adam Shajnfeld and Adithya Gangidi and Adolfo Victoria and Ahuva Goldstand and Ajay Menon and Ajay Sharma and Alex Boesenberg and Alexei Baevski and Allie Feinstein and Amanda Kallet and Amit Sangani and Amos Teo and Anam Yunus and Andrei Lupu and Andres Alvarado and Andrew Caples and Andrew Gu and Andrew Ho and Andrew Poulton and Andrew Ryan and Ankit Ramchandani and Annie Dong and Annie Franco and Anuj Goyal and Aparajita Saraf and Arkabandhu Chowdhury and Ashley Gabriel and Ashwin Bharambe and Assaf Eisenman and Azadeh Yazdan and Beau James and Ben Maurer and Benjamin Leonhardi and Bernie Huang and Beth Loyd and Beto De Paola and Bhargavi Paranjape and Bing Liu and Bo Wu and Boyu Ni and Braden Hancock and Bram Wasti and Brandon Spence and Brani Stojkovic and Brian Gamido and Britt Montalvo and Carl Parker and Carly Burton and Catalina Mejia and Ce Liu and Changhan Wang and Changkyu Kim and Chao Zhou and Chester Hu and Ching-Hsiang Chu and Chris Cai and Chris Tindal and Christoph Feichtenhofer and Cynthia Gao and Damon Civin and Dana Beaty and Daniel Kreymer and Daniel Li and David Adkins and David Xu and Davide Testuggine and Delia David and Devi Parikh and Diana Liskovich and Didem Foss and Dingkang Wang and Duc Le and Dustin Holland and Edward Dowling and Eissa Jamil and Elaine Montgomery and Eleonora Presani and Emily Hahn and Emily Wood and Eric-Tuan Le and Erik Brinkman and Esteban Arcaute and Evan Dunbar and Evan Smothers and Fei Sun and Felix Kreuk and Feng Tian and Filippos Kokkinos and Firat Ozgenel and Francesco Caggioni and Frank Kanayet and Frank Seide and Gabriela Medina Florez and Gabriella Schwarz and Gada Badeer and Georgia Swee and Gil Halpern and Grant Herman and Grigory Sizov and Guangyi and Zhang and Guna Lakshminarayanan and Hakan Inan and Hamid Shojanazeri and Han Zou and Hannah Wang and Hanwen Zha and Haroun Habeeb and Harrison Rudolph and Helen Suk and Henry Aspegren and Hunter Goldman and Hongyuan Zhan and Ibrahim Damlaj and Igor Molybog and Igor Tufanov and Ilias Leontiadis and Irina-Elena Veliche and Itai Gat and Jake Weissman and James Geboski and James Kohli and Janice Lam and Japhet Asher and Jean-Baptiste Gaya and Jeff Marcus and Jeff Tang and Jennifer Chan and Jenny Zhen and Jeremy Reizenstein and Jeremy Teboul and Jessica Zhong and Jian Jin and Jingyi Yang and Joe Cummings and Jon Carvill and Jon Shepard and Jonathan McPhie and Jonathan Torres and Josh Ginsburg and Junjie Wang and Kai Wu and Kam Hou U and Karan Saxena and Kartikay Khandelwal and Katayoun Zand and Kathy Matosich and Kaushik Veeraraghavan and Kelly Michelena and Keqian Li and Kiran Jagadeesh and Kun Huang and Kunal Chawla and Kyle Huang and Lailin Chen and Lakshya Garg and Lavender A and Leandro Silva and Lee Bell and Lei Zhang and Liangpeng Guo and Licheng Yu and Liron Moshkovich and Luca Wehrstedt and Madian Khabsa and Manav Avalani and Manish Bhatt and Martynas Mankus and Matan Hasson and Matthew Lennie and Matthias Reso and Maxim Groshev and Maxim Naumov and Maya Lathi and Meghan Keneally and Miao Liu and Michael L. Seltzer and Michal Valko and Michelle Restrepo and Mihir Patel and Mik Vyatskov and Mikayel Samvelyan and Mike Clark and Mike Macey and Mike Wang and Miquel Jubert Hermoso and Mo Metanat and Mohammad Rastegari and Munish Bansal and Nandhini Santhanam and Natascha Parks and Natasha White and Navyata Bawa and Nayan Singhal and Nick Egebo and Nicolas Usunier and Nikhil Mehta and Nikolay Pavlovich Laptev and Ning Dong and Norman Cheng and Oleg Chernoguz and Olivia Hart and Omkar Salpekar and Ozlem Kalinli and Parkin Kent and Parth Parekh and Paul Saab and Pavan Balaji and Pedro Rittner and Philip Bontrager and Pierre Roux and Piotr Dollar and Polina Zvyagina and Prashant Ratanchandani and Pritish Yuvraj and Qian Liang and Rachad Alao and Rachel Rodriguez and Rafi Ayub and Raghotham Murthy and Raghu Nayani and Rahul Mitra and Rangaprabhu Parthasarathy and Raymond Li and Rebekkah Hogan and Robin Battey and Rocky Wang and Russ Howes and Ruty Rinott and Sachin Mehta and Sachin Siby and Sai Jayesh Bondu and Samyak Datta and Sara Chugh and Sara Hunt and Sargun Dhillon and Sasha Sidorov and Satadru Pan and Saurabh Mahajan and Saurabh Verma and Seiji Yamamoto and Sharadh Ramaswamy and Shaun Lindsay and Shaun Lindsay and Sheng Feng and Shenghao Lin and Shengxin Cindy Zha and Shishir Patil and Shiva Shankar and Shuqiang Zhang and Shuqiang Zhang and Sinong Wang and Sneha Agarwal and Soji Sajuyigbe and Soumith Chintala and Stephanie Max and Stephen Chen and Steve Kehoe and Steve Satterfield and Sudarshan Govindaprasad and Sumit Gupta and Summer Deng and Sungmin Cho and Sunny Virk and Suraj Subramanian and Sy Choudhury and Sydney Goldman and Tal Remez and Tamar Glaser and Tamara Best and Thilo Koehler and Thomas Robinson and Tianhe Li and Tianjun Zhang and Tim Matthews and Timothy Chou and Tzook Shaked and Varun Vontimitta and Victoria Ajayi and Victoria Montanez and Vijai Mohan and Vinay Satish Kumar and Vishal Mangla and Vlad Ionescu and Vlad Poenaru and Vlad Tiberiu Mihailescu and Vladimir Ivanov and Wei Li and Wenchen Wang and Wenwen Jiang and Wes Bouaziz and Will Constable and Xiaocheng Tang and Xiaojian Wu and Xiaolan Wang and Xilun Wu and Xinbo Gao and Yaniv Kleinman and Yanjun Chen and Ye Hu and Ye Jia and Ye Qi and Yenda Li and Yilin Zhang and Ying Zhang and Yossi Adi and Youngjin Nam and Yu and Wang and Yu Zhao and Yuchen Hao and Yundi Qian and Yunlu Li and Yuzi He and Zach Rait and Zachary DeVito and Zef Rosnbrick and Zhaoduo Wen and Zhenyu Yang and Zhiwei Zhao and Zhiyu Ma},
      year={2024},
      eprint={2407.21783},
      archivePrefix={arXiv},
      primaryClass={cs.AI},
      url={https://arxiv.org/abs/2407.21783}, 
}

@misc{li2023textbooksneediiphi15,
      title={Textbooks Are All You Need II: phi-1.5 technical report}, 
      author={Yuanzhi Li and Sébastien Bubeck and Ronen Eldan and Allie Del Giorno and Suriya Gunasekar and Yin Tat Lee},
      year={2023},
      eprint={2309.05463},
      archivePrefix={arXiv},
      primaryClass={cs.CL},
      url={https://arxiv.org/abs/2309.05463}, 
}

@article{javaheripi2023phi,
  title={Phi-2: The surprising power of small language models},
  author={Javaheripi, Mojan and Bubeck, S{\'e}bastien and Abdin, Marah and Aneja, Jyoti and César Teodoro Mendes, Caio and Chen, Weizhu and Del Giorno, Allie and Eldan, Ronen and Gopi, Sivakanth and Gunasekar, Suriya and Kauffmann, Piero and Lee, Yin Tat and Li, Yuanzhi and Nguyen,  Anh  and Rosa, Gustavo de and Saarikivi, Olli and Salim, Adil and Shah, Shital and Santacroce, Michael and Singh Behl, Harkirat and Taumann Kalai, Adam and Wang, Xin and Ward, Rachel and Witte, Philipp and Zhang, Cyril and Zhang, Yi},
  journal={Microsoft Research Blog},
  year={2023}
}

@article{hanna2023does,
  title={How does GPT-2 compute greater-than?: Interpreting mathematical abilities in a pre-trained language model},
  author={Hanna, Michael and Liu, Ollie and Variengien, Alexandre},
  journal={Advances in Neural Information Processing Systems},
  volume={36},
  pages={76033--76060},
  year={2023}
}

@article{mathwin2023identifying,
  title={Identifying a preliminary circuit for predicting gendered pronouns in gpt-2 small},
  author={Mathwin, Chris and Corlouer, Guillaume and Kran, Esben and Barez, Fazl and Nanda, Neel},
  journal={URL: https://itch. io/jam/mechint/rate/1889871},
  year={2023}
}

@inproceedings{
hsu2025efficient,
title={Efficient Automated Circuit Discovery in Transformers using Contextual Decomposition},
author={Aliyah R. Hsu and Georgia Zhou and Yeshwanth Cherapanamjeri and Yaxuan Huang and Anobel Odisho and Peter R. Carroll and Bin Yu},
booktitle={The Thirteenth International Conference on Learning Representations},
year={2025},
url={https://openreview.net/forum?id=41HlN8XYM5}
}

@inproceedings{hanna2024have,
title={Have Faith in Faithfulness: Going Beyond Circuit Overlap When Finding Model Mechanisms},
author={Michael Hanna and Sandro Pezzelle and Yonatan Belinkov},
booktitle={First Conference on Language Modeling},
year={2024},
url={https://openreview.net/forum?id=TZ0CCGDcuT}
}

@misc{nanda2023attribution,
  author       = {Neel Nanda and Chris Olah and Catherine Olsson and Nelson Elhage and Hume Tristan},
  title        = {Attribution Patching: Activation Patching at Industrial Scale},
  year         = {2023},
  url          = {https://www.neelnanda.io/mechanistic-interpretability/attribution-patching},
}

@misc{nanda2022transformerlens,
    title = {TransformerLens},
    author = {Neel Nanda and Joseph Bloom},
    year = {2022},
    howpublished = {\url{https://github.com/TransformerLensOrg/TransformerLens}},
}

@inproceedings{paszke2019pytorch,
  title={PyTorch: An imperative style, high-performance deep learning library},
  author={Paszke, Adam and Gross, Sam and Massa, Francisco and Lerer, Adam and Bradbury, James and Chanan, Gregory and Killeen, Trevor and Lin, Zeming and Tompson, Estelle and Desmaison, Laurens and Kopf, Alban and Yang, Edward and DeVito, Zachary and Raison, Martin and Tejani, Alykhan and Chilimbi, Sasank and Prabhakaran, Benoit},
  booktitle={Advances in neural information processing systems},
  pages={8024--8035},
  year={2019}
}

@misc{sung2021trainingneuralnetworksfixed,
      title={Training Neural Networks with Fixed Sparse Masks}, 
      author={Yi-Lin Sung and Varun Nair and Colin Raffel},
      year={2021},
      eprint={2111.09839},
      archivePrefix={arXiv},
      primaryClass={cs.LG},
      url={https://arxiv.org/abs/2111.09839}, 
}

@inproceedings{joshi-etal-2025-calibration,
    title = "Calibration Across Layers: Understanding Calibration Evolution in {LLM}s",
    author = "Joshi, Abhinav  and
      Ahmad, Areeb  and
      Modi, Ashutosh",
    editor = "Christodoulopoulos, Christos  and
      Chakraborty, Tanmoy  and
      Rose, Carolyn  and
      Peng, Violet",
    booktitle = "Proceedings of the 2025 Conference on Empirical Methods in Natural Language Processing",
    month = nov,
    year = "2025",
    address = "Suzhou, China",
    publisher = "Association for Computational Linguistics",
    url = "https://aclanthology.org/2025.emnlp-main.742/",
    doi = "10.18653/v1/2025.emnlp-main.742",
    pages = "14697--14725",
    ISBN = "979-8-89176-332-6"
}

@inproceedings{joshi-etal-2025-towards,
    title = "Towards Quantifying Commonsense Reasoning with Mechanistic Insights",
    author = "Joshi, Abhinav  and
      Ahmad, Areeb  and
      Shukla, Divyaksh  and
      Modi, Ashutosh",
    editor = "Chiruzzo, Luis  and
      Ritter, Alan  and
      Wang, Lu",
    booktitle = "Proceedings of the 2025 Conference of the Nations of the Americas Chapter of the Association for Computational Linguistics: Human Language Technologies (Volume 1: Long Papers)",
    month = apr,
    year = "2025",
    address = "Albuquerque, New Mexico",
    publisher = "Association for Computational Linguistics",
    url = "https://aclanthology.org/2025.naacl-long.487/",
    doi = "10.18653/v1/2025.naacl-long.487",
    pages = "9633--9660",
    ISBN = "979-8-89176-189-6"
}

@inproceedings{
joshi2024cold,
title={{COLD}: Causal reasOning in cLosed Daily activities},
author={Abhinav Joshi and Areeb Ahmad and Ashutosh Modi},
booktitle={The Thirty-eighth Annual Conference on Neural Information Processing Systems},
year={2024},
url={https://openreview.net/forum?id=7Mo1NOosNT}
}
\bibliographystyle{plainnat}

\clearpage
\newpage
\clearpage
\newpage

\section*{Appendix}

\appendix

%%%%%%%%%%%%%%%%%%%%%%%%%%%

\hypersetup{linkcolor=blue}

% \titlecontents{section}[18pt]{% % \vspace{0.05em}}{\contentslabel{1.5em}}{}
% {\titlerule*[0.5pc]{.}\contentspage} % Set the formatting for appendix sections in the table of contents
% \titlecontents{table}[0pt]{% % \vspace{0.05em}}{\contentslabel{1em}}{}
% {\titlerule*[0.5pc]{.}\contentspage} % Set the formatting for appendix tables in the list of tables

\startcontents[appendix] % Start the table of contents for the appendix
\section*{Table of Contents} % Title for the appendix table of contents
%\addcontentsline{toc}{section}{Table of Contents} % Add the appendix table of contents to the main table of contents
\printcontents[appendix]{section}{0}{\setcounter{tocdepth}{4}} % Print the table of contents for the appendix

\startlist[appendix]{lot} % Start the list of tables for the appendix
\section*{List of Tables} % Title for the appendix list of tables
%\addcontentsline{lot}{section}{List of Tables} % Add the appendix list of tables to the main list of tables
\printlist[appendix]{lot}{}{\setcounter{tocdepth}{2}} % Print the list of tables for the appendix

\startlist[appendix]{lof} % Start the list of figures for the appendix
\section*{List of Figures} % Title for the appendix list of tables
%\addcontentsline{lot}{section}{List of Tables} % Add the appendix list of tables to the main list of tables
\printlist[appendix]{lof}{}{\setcounter{tocdepth}{2}}

%\listofappendixfigures
\newpage

%%%%%%%%%%%%%%%%%%%%%%%%%%%
%%%%%%%%%%%%%%%%%%%%%%%
\section{Details of Datasets} \label{app:sec-real-dataset}
This section provides details about the datasets used in our experiments. While the Indirect Object Identification (IOI) task is our primary benchmark due to its strong alignment with our goals in mechanistic interpretability, we also include two additional datasets, Greater-Than \citep{hanna2023does} and Gender Pronoun \citep{mathwin2023identifying}, to test the generalizability of our method.

\subsection{Indirect Object Identification (IOI)}
\label{app:dataset-ioi}

The IOI task~\citep{wang2022interpretabilitywildcircuitindirect} is a synthetic benchmark designed to study a language model’s ability to resolve coreference between proper names in complex syntactic constructions. Each prompt contains two names in an introductory clause, followed by a clause in which one name acts as the subject. The model must complete the sentence with the indirect object:
\begin{quote}
    \textit{``When Mary and John went to the store, John gave a drink to''} $\rightarrow$ \textbf{Mary}
\end{quote}

The task follows a known and interpretable algorithm, i.e., predict the name that is \textbf{not} the subject of the last clause. Its controlled structure and clear ground truth make it ideal for circuit-level analysis and interpreting internal representations of models like GPT-2.

\subsection{Gender Pronoun Task}
\label{app:dataset-gender}

This task evaluates pronoun resolution in socially relevant contexts. Each example is a declarative sentence about a named person followed by a tag question. The goal is to complete the sentence with the correct gendered pronoun:

\begin{quote}
\textit{``So David is a really great friend, isn’t''} $\rightarrow$ \textbf{he} \\
\textit{``So Mary is a very good athlete, isn’t''} $\rightarrow$ \textbf{she}
\end{quote}

This dataset complements IOI by testing the model’s ability to resolve pronouns and capture gender semantics. It also provides insight into whether our method can disentangle social biases embedded in model representations.

\subsection{Greater-Than Task}
\label{app:dataset-greaterthan}

The Greater-Than task \citep{hanna2023does} targets numerical reasoning by prompting the model with a sentence involving two years. The goal is to predict a valid end year that is greater than the start year, using a textual format:

\begin{quote}
\textit{``The treaty lasted from the year 1314 to the year 13''} $\rightarrow$ \textbf{28}
\end{quote}

The dataset is automatically generated using 120 nouns representing temporal events (sourced from FrameNet), with years sampled from the 11th to 17th centuries. Special care is taken to avoid multi-token year completions and boundary cases, ensuring all completions are single-token and meaningful under GPT-2’s tokenizing scheme.

\section{Additional Results, Discussion and Future Directions} \label{app:additional-results-section}
In this appendix section, we provide extended results, analysis, and exploratory observations that complement the findings in the main paper. We provide further support for the interpretability of low-rank directions across multiple tasks, highlight composite functional patterns, and outline promising directions for future work.

\subsection{Compute Resources}
\label{app-sec:compute-resources}

All experiments were conducted on a single NVIDIA A40 GPU with 48 GB of VRAM. Our implementation is based on PyTorch 2.1 \citep{paszke2019pytorch} and the HuggingFace Transformers library (v4.35), leveraging its integration with pre-trained models and tokenizer utilities. We utilize the GPT-2 small model (124M parameters), with all weights frozen during our experiments to ensure the integrity of the underlying representations. For systematic access to internal model activations and component-wise analysis, we employ the \texttt{TransformerLens} library~\citep{nanda2022transformerlens}, which provides fine-grained control over the transformer’s intermediate computations, enabling singular value decomposition and mask-based interventions at the component level. All training for our mask optimization and probing modules was conducted using batches of IOI or Greater-Than task prompts, with early stopping based on validation loss to ensure efficiency and generalizability. The details to the Dataset splits and Hyperparameters are provided in Table \ref{tab:dataset_splits} and Table \ref{tab:hyperparameters}, respectively.
Table~\ref{tab:dataset_splits} outlines the dataset splits used across different tasks, Indirect Object Identification (IOI), Greater-Than (GT), and Generalized Preposition (GP). We use relatively small training sets to textitasize the interpretability and efficiency of our method, while validation and test sets are significantly larger to ensure robust generalization. 
The training procedure for our low-rank component masks and probing modules was consistent across tasks, with hyperparameters detailed in Table~\ref{tab:hyperparameters}. We optimize using AdamW with moderate weight decay and L1 regularization to encourage sparsity in learned masks. Training is conducted for a maximum of 150 epochs with early stopping based on validation performance to avoid overfitting. The relatively small batch size and learning rate were empirically found to balance convergence speed with stability.

\begin{table}[t]
    \centering
    \caption{Dataset splits for the three tasks used in our experiments, IOI (Indirect Object Identification), GT (Gender Type), and GP (Gender Pronouns), indicating the number of examples allocated to the training, validation, and test sets.}
    \begin{tabular}{l|c|c|c}
        \toprule
        \textbf{Task} & \textbf{Train} & \textbf{Validation} & \textbf{Test} \\
        \midrule
        IOI & 1k & 200 & 1k \\
        % \hline
        GT & 2k & 500 & 2k \\
        % \hline
        GP & 1k & 155 & 307 \\
        \bottomrule
    \end{tabular}
    \label{tab:dataset_splits}
\end{table}

\begin{table}[t]
    \caption{Hyperparameters used for training the linear probes and other learned components in our experiments. We report batch size, number of epochs, optimization parameters (learning rate and weight decay), and the coefficient of the L1 regularization term.}
    \label{tab:hyperparameters}
    \centering
    \begin{tabular}{ll}
        \toprule
        \textbf{Hyperparameter} & \textbf{Value} \\
        % \midrule
        % \multicolumn{2}{l}{\textit{Model Configuration}} \\
        % \midrule
        % Axis Padding Length & 64 \\
        % \midrule
        % \multicolumn{2}{l}{\textit{Training Configuration}} \\
        \midrule
        Batch Size & 64 \\
        Number of Epochs & 15 \\
        % Device & CUDA \\
        Learning Rate & $1.0 \times 10^{-2}$ \\
        Weight Decay & $1.0 \times 10^{-9}$ \\
        L1 Regularization Weight & $1.5 \times 10^{-4}$ \\
        \bottomrule
    \end{tabular}
\end{table}

\subsection{Inherently formed Logit Directions}
\label{app:Logit-directions}

Beyond component-level decomposition, our analysis reveals that in transformer layers there exist a set of \textit{inherent, fixed directions} in logit space, which we term as \textit{"Logit Receptors"} (so named because they act as fixed receptive directions that {deterministically} modulate specific token logits). These are directions that correspond to stable vocabulary preferences learned during training and that operate as modulators for specific tokens. These directions emerge naturally from the singular value decomposition of the augmented OV projection, \(\mathbf{W}_{\mathrm{aug}}^{(\mathrm{OV})}\), which writes the attention output into the residual stream.

\paragraph{From Singular Vectors to Logit Receptors.}
Let \(\nu_{lh}\in\mathbb{R}^{(1+d_{\text{model}})}\) denote the attention-weighted (and {one}-augmented, to compensate bias) value vector for layer \(l\), head \(h\), where we augment the context with a constant 1 so that \(\nu_{lh}^\top=[1, \sum_j \alpha_ij x_j]\) in the notation above($\alpha$ is attention distribution and $x_j$ is post-layernorm residual stream representation). Decomposing the corresponding OV projection by singular value decomposition gives
\[
\mathbf{W}_{\mathrm{aug}}^{(\mathrm{OV})} = U_{lh}\,\Sigma_{lh}\,V_{lh}^\top .
\]
The output, in equation \eqref{eq:ov-circuit-output}, written by head \(h\) in layer \(l\) {in the residual stream} can then be expressed as 
\[
y_{lh}
= \nu_{lh}^\top\,\mathbf{W}_{\mathrm{aug}}^{(\mathrm{OV})}
= \sum_k \sigma_{lhk}\,(\nu_{lh}^\top u_{lhk})\,v_{lhk}^\top,
\]
where each term corresponds to one singular direction \(k\) with singular value \(\sigma_{lhk}\), left singular vector \(u_{lhk}\), and right singular vector \(v_{lhk}\). The scalar term \(\nu_{lh}^{\top} u_{lhk}\) represents an input-dependent coefficient that {steers} a fixed direction in residual space.

When each right-singular vector \(v_{lhk}\) is projected through the unembedding matrix \(W_U\), we obtain a scaled vocabulary-length vector
\[
y_{lhk}
= (\sigma_{lhk}\,(\nu_{lh}^{\top}u_{lhk}))\,v_{lhk}^\top W_U,
\]
which corresponds to a {fixed direction} (pointing towards a specific set of tokens) in token-logit space. Importantly, the term \(v_{lhk}^\top W_U\) depends only on the learned weights; in other words, it is a constant Logit Receptor independent of any specific input. The only input-dependent term is its {scalar} activation, \(\nu_{lh}^{\top}u_{lhk}\), which modulates how strongly the model calls upon that logit receptor at a given step.

\begin{figure}
    \centering
    \includegraphics[width=0.75\linewidth]{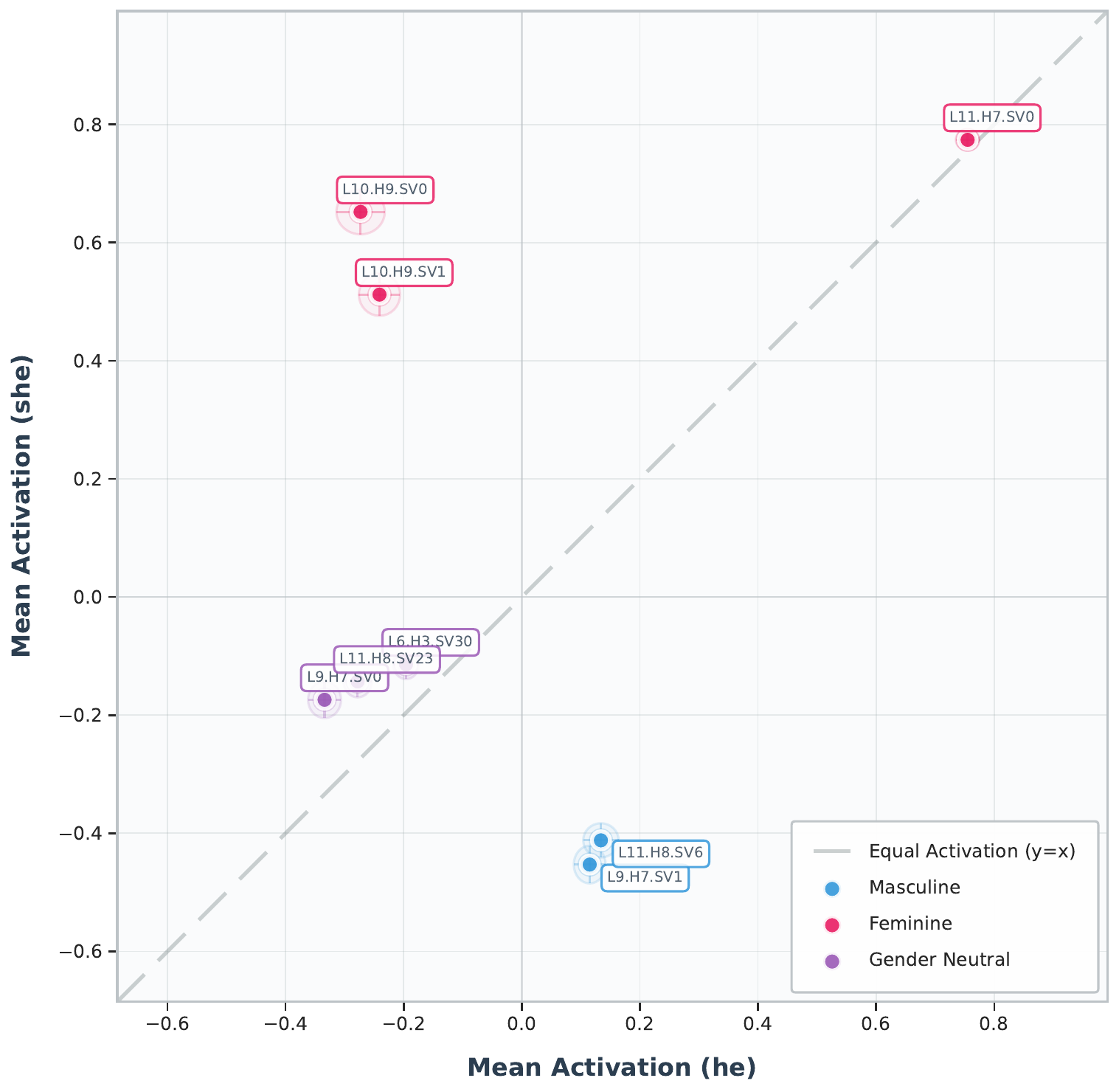}
    \caption{Mean activation of gender-related directions conditioned on \textit{Masculine} versus \textit{Feminine} prompt context. The x-axis plots the mean activation $\mathbb{E}[\nu^{\top}u \mid \text{prompt context=he}]$ and the y-axis plots mean $\mathbb{E}[\nu^{\top}u \mid \text{prompt context=she}]$. Error bars show one standard deviation. 
    % Points are colored by scheme: Masculine (blue), Feminine (red), and Gender Neutral (gray). 
    The dashed diagonal line represents $y=x$, where activations for both pronouns would be equal.}
\label{fig:gender-activation-plot}
\end{figure}

\paragraph{Interpretable Logit Receptors.}
Empirically, we found that some of these fixed logit receptors have clear linguistic meaning. In the Gender Pronoun Resolution (GP) task, for instance, certain \(v_{lhk}^\top W_U\) vectors are more inclined towards gendered pronouns. Their corresponding activations \(\nu_{lh}^{\top}u_{lhk}\) are systematically higher for examples of one gender than the other, forming a mechanistic pattern, i.e., the model maintains distinct “she” and “he” receptors in logit space and routes activation selectively between them depending on context.

This perspective reveals a fresh way of understanding concepts present inherently in the internal structure of the model, even before considering the mask-learning step; the network already contains a compact basis of hard-coded logit receptors that modulate the vocabulary space. These receptors define semantic axes that downstream layers can selectively activate to express context-sensitive meaning. Learning directional masks (via Algorithm \ref{alg:maskoptimization}) further clarifies which of these fixed logit receptors are actually used by the task. Directions with high mask weights may correspond to those carrying discriminative information for the task, while others are effectively pruned. Moreover, it also reflects that some of the inherent structures are not used by a specific dataset, and the discovery of components is highly dependent on the used dataset, which may not contain all the universal inherent directions present in a model.

A notable example is \texttt{L11.H7.SV0}, where despite having the largest singular value (\(\sigma = 22.07\)) and a vocabulary projection strongly associated with feminine tokens, this direction receives an almost-zero mask. Inspection reveals that its activation \(\nu_{lh}^{\top}u_{lhk}\) (a scalar modulating the logit receptor) is nearly identical for male and female examples, essentially encoding variance but no discriminative signal (also see Figure~\ref{fig:gender-activation-plot}). In contrast, directions such as \texttt{L10.H9.SV0} show pronounced activation differences across gender contexts and are thus retained. 
Table~\ref{tab:gender-directions} summarizes the salient discovered directions, including their mask weights, singular values, top-vocabulary tokens, and conditional activation statistics. Together, these results suggest that the model’s internal computation operates through a set of inherently learned logit receptors, fixed basis vectors whose selective activation underlies task behavior and provides a fine-grained, mechanistic bridge between representation and prediction.

\begin{table*}[t]
\centering
\caption{Selected gender- and number-related OV singular directions.
Columns show the learned mask, singular value $\sigma$, an excerpt of top tokens from $v^\top W_U$, mean activation $\nu^{\top}u$ (mean $\pm$ std) conditioned on the pronoun based on prompt context($p_c$), and the activation difference (he minus she).}
\footnotesize
\renewcommand{\arraystretch}{1.05}
\setlength{\tabcolsep}{6pt}
% Define a ragged-right X column for long token lists
\newcolumntype{Y}{>{\raggedright\arraybackslash}X}

\begin{tabularx}{\textwidth}{@{}l c c Y c c c@{}}
\toprule
Direction & Mask & $\sigma$ & Top tokens (excerpt) &
$\mathbb{E}[\nu^{\top}u\mid\text{$p_c$=he}]$ & $\mathbb{E}[\nu^{\top}u\mid\text{$p_c$=she}]$ & Diff \\
\midrule

%\definecolor{masculinecolor}{rgb}{0.8, 0.9, 1.0} % Light Blue
%\definecolor{femininecolor}{rgb}{1.0, 0.9, 0.9} % Light Pink
%\definecolor{neutralcolor}
% --- Masculine ---
%\rowcolor{blue!20}
\rowcolor{masculinecolor} 
\texttt{L9.H7.SV1} & 1.00 & 8.87 & His, his, He, he, himself & $+0.115\pm0.027$ & $-0.453\pm0.032$ & $+0.568$ \\[4pt]
%\rowcolor{blue!20}
%\rowcolor{masculinecolor} 
\texttt{L11.H8.SV6} & 1.00 & 6.52 & his, His, him, He, he & $+0.134\pm0.030$ & $-0.412\pm0.029$ & $+0.546$ \\[4pt]
\midrule

% --- Feminine ---
%\rowcolor{red!20}
\rowcolor{femininecolor} 
\texttt{L10.H9.SV0} & 1.00 & 9.15 & her, she, She, herself, hers & $-0.273\pm0.041$ & $+0.652\pm0.038$ & $-0.925$ \\[4pt]
%\rowcolor{red!20}
\texttt{L11.H7.SV0} & 1.3e-5 & 22.07 & herself, Her, her, hers, She & $+0.755\pm0.020$ & $+0.774\pm0.018$ & $-0.019$ \\[4pt]
%\rowcolor{red!20}
\texttt{L10.H9.SV1} & 0.82 & 7.87 & her, she, herself, She, she & $-0.241\pm0.035$ & $+0.512\pm0.036$ & $-0.753$ \\[6pt]
\midrule

% --- They/Them (Neutral) ---
%\rowcolor{gray!20}
\rowcolor{neutralcolor}
\texttt{L9.H7.SV0} & 1.00 & 9.23 & their, Their, they, their & $-0.334\pm0.028$ & $-0.174\pm0.031$ & $-0.160$ \\[4pt]
%\rowcolor{gray!20}
\texttt{L6.H3.SV30} & 1.00 & 4.64 & they, They, Them, Their & $-0.196\pm0.022$ & $-0.114\pm0.025$ & $-0.082$ \\[4pt]
%\rowcolor{gray!20}
\texttt{L11.H8.SV23} & 0.87 & 4.32 & their, Their, THEY, THEIR, They & $-0.278\pm0.024$ & $-0.143\pm0.027$ & $-0.135$ \\
\bottomrule
\end{tabularx}
\label{tab:gender-directions}
\end{table*}

This decomposition helps provide an understanding of how the model performs token-level decisions. Each {unembedding projection of} right-singular direction {i.e. }\(v_{lhk}^\top W_U\) acts as a \textit{fixed}{ logit receptor} controlling the prediction vocabulary space, essentially representing a stable axis associated with a specific token outcome.
During inference, the model dynamically steers these receptors through the input-dependent activation coefficients \(\nu_{lh}^{\top}u_{lhk}\). The masks we learn over these singular directions identify which of these receptors the model actually employs, i.e., directions that combine both {capacity} (large singular value) and {discriminative alignment} (systematic activation shifts across labels) receive high mask weights, while those that contribute variance but no discriminative signal are suppressed during optimization.  
% \Areeb
{As highlighted in Figure~\ref{fig:gender-activation-plot}, the activations \(\nu_{lh}^{\top}u_{lhk}\) form linearly separable clusters for masculine and feminine contexts; directions lacking a decisive separation receive negligible mask weights, indicating that the masking learns to ignore non-discriminative directions.} For example, direction \texttt{L11.H7.SV0}, despite its large singular value (\(\sigma=22.07\)) and a clearly feminine token projection, receives a near-zero mask because its activation distribution is nearly identical across male and female examples.

We further ask what happens if we intervene in these directions. While the above analysis reveals interpretable correlations between singular directions and task behavior, it does not establish whether these directions are \textit{causally responsible} for model predictions.
To support our central hypothesis, that transformers distribute computation along distinct, low-rank subfunctions, we require causal evidence that manipulating these subspaces directly alters model output in predictable ways. If a small number of identified singular directions are genuinely responsible for a behavior, then systematically altering them should induce controlled, interpretable changes in the model’s output logits, while leaving other computations intact.

\paragraph{Scalar-based Causal Ablation of OV Directions.}
We therefore perform a series of \textit{Scalar-based counterfactual ablations} designed to test whether the discovered singular directions in the OV projection causally steer gender pronoun prediction. In each ablation, we swap the activations of gender-sensitive singular directions (those with high learned mask values) with the empirically observed mean activations from the opposite gender distribution.
For instance, the masculine-sensitive direction \texttt{L9.H7.SV1} typically exhibits mean activations of \(+0.115\) for “he” prompts and \(-0.453\) for “she” prompts; the intervention exchanges these values, effectively inserting counterfactual gender evidence while keeping all other activations fixed.

Overall, we conduct four controlled experiments: 1) swapping all gender directions for masculine prompts, 2) swapping all gender directions for feminine prompts, 3) swapping only masculine directions for masculine prompts, and 4) swapping only feminine directions for feminine prompts.
Each condition measures how average pronoun-logit differences respond to these targeted perturbations, thereby quantifying the causal influence of individual low-rank subspaces.

Formally, for a given singular direction \(i\), the intervention replaces the natural activation \((\nu^{\top}u_i)\) with the opposite-gender mean activation \(a'_i\):
\[
\Delta R = (a'_i - \nu^{\top}u_i)\,\sigma_i\,v_i^\top .
\]
% \Areeb
{adding this re-scaled vector to the final residual stream vector directly intervenes in the contribution of specific interpretable low-rank features, allowing precise causal editing of gender-specific behavior. Algorithm \ref{alg:range-swap} illustrates this process in more detail.}
This precisely modifies the residual-stream contribution of selected interpretable features while preserving the model’s overall structure, enabling rigorous, scalar-controlled causal testing.

For empirical validation, we consider instances from the GP (Gender Pronoun Resolution) task and segregate prompts into distinct masculine ($p_{\text{male}}$) and feminine ($p_{\text{female}}$) contexts and pass them separately to the model and record the scalar mean corresponding to specific tokens ($
\mu_{g} = 
a'_{i,g}
    = \frac{1}{|p_g|}
      \sum_{x \in p_g} \nu^\top u_i,\quad
g \in \{\mathrm{male},\, \mathrm{female}\}
$). This helps provide mean values for each considered context, and for all the pronoun directions $i\in\{\texttt{L9.H7.SV1}\textbf{, } \texttt{L11.H8.SV6}\textbf{, }.......\}$, which are later used in Algorithm \ref{alg:range-swap} to perform intervention and observe the effect of modifying a logit receptor. (also see Figure \ref{fig:intervention-logit-receptor}, that illustrates the intervention performed in Algorithm \ref{alg:range-swap})

\begin{algorithm}[t]
\caption{Intervention for OV Singular Directions}
\label{alg:range-swap}
\footnotesize
\begin{algorithmic}[1]
\Require Model $\mathcal{M}$, SVD circuit cache $\mathcal{C}$, data loader $\mathcal{D}$,
directions $\mathcal{S}$ (each with $(\ell, h, i, \mu_{\text{he}}, \mu_{\text{she}})$), and target gender $g\in\{\text{he},\text{she}\}$
\Ensure Summary statistics: mean logit difference, variance, etc.
\vspace{0.5em}

\State Initialize $\text{results}\leftarrow\{\text{correct},\text{logit\_diff},\text{they\_logit}\}$

\For{each batch $(x,y)$ in $\mathcal{D}$}
  \State Run model with cache: $(\text{logits}, \text{cache}) \gets \mathcal{M}.\text{run\_with\_cache}(x)$
  \State Extract final-token index $t^\ast$
  % \State Determine target mask $\mathbb{1}_{g}(y)$
  \State Initialize $\Delta R \gets 0$

  \For{each $(\ell,h,i,\mu_{\text{he}},\mu_{\text{she}})\in\mathcal{S}$}
      \State Retrieve SVD components $U,S,V \gets \mathcal{C}[\ell,h]$
      \State Select $(u_i,v_i,\sigma_i)\gets(U[:,i],V[:,i],S[i])$
      \State Compute attention-weighted context $\nu=\,[1, \sum \alpha x]$
      \State Current activation $a_i=(\nu^{\top} u_i)$
      
      \State Target activation $a'_i \gets
        \begin{cases}
           \mu_{\text{she}}, & g=\text{he}\\
           \mu_{\text{he}}, & g=\text{she}
        \end{cases}$
      \State $\Delta a_i \gets (a'_i-a_i)$
      \State $\sigma_i\leftarrow \sigma_{scale}\times\sigma_i$
      \State $\Delta R \mathrel{+}= \Delta a_i\, \sigma_i\, v_i^{\top}$
  \EndFor

  \State Add $\Delta R$ to final residual stream at $t^\ast$
  \State Compute new logits $Z' = \mathrm{LayerNorm}(\text{residual}+\Delta R)W_U + b_U$
  % \State Extract pronoun logits $Z'_{\text{he}},Z'_{\text{she}},Z'_{\text{they}}$
  % \State $\text{logit\_diff}\gets (Z'_{\text{he}}-Z'_{\text{she}})$ or vice versa i.e. 
  % \State Append to $\text{results}$
\EndFor

% \State Compute summary:
% $\text{mean}(\text{logit\_diff})$, $\text{std}(\text{logit\_diff})$
% \State \Return summary
\end{algorithmic}
\end{algorithm}

\begin{figure}
    \centering\
    
    \includegraphics[width=0.95\linewidth]{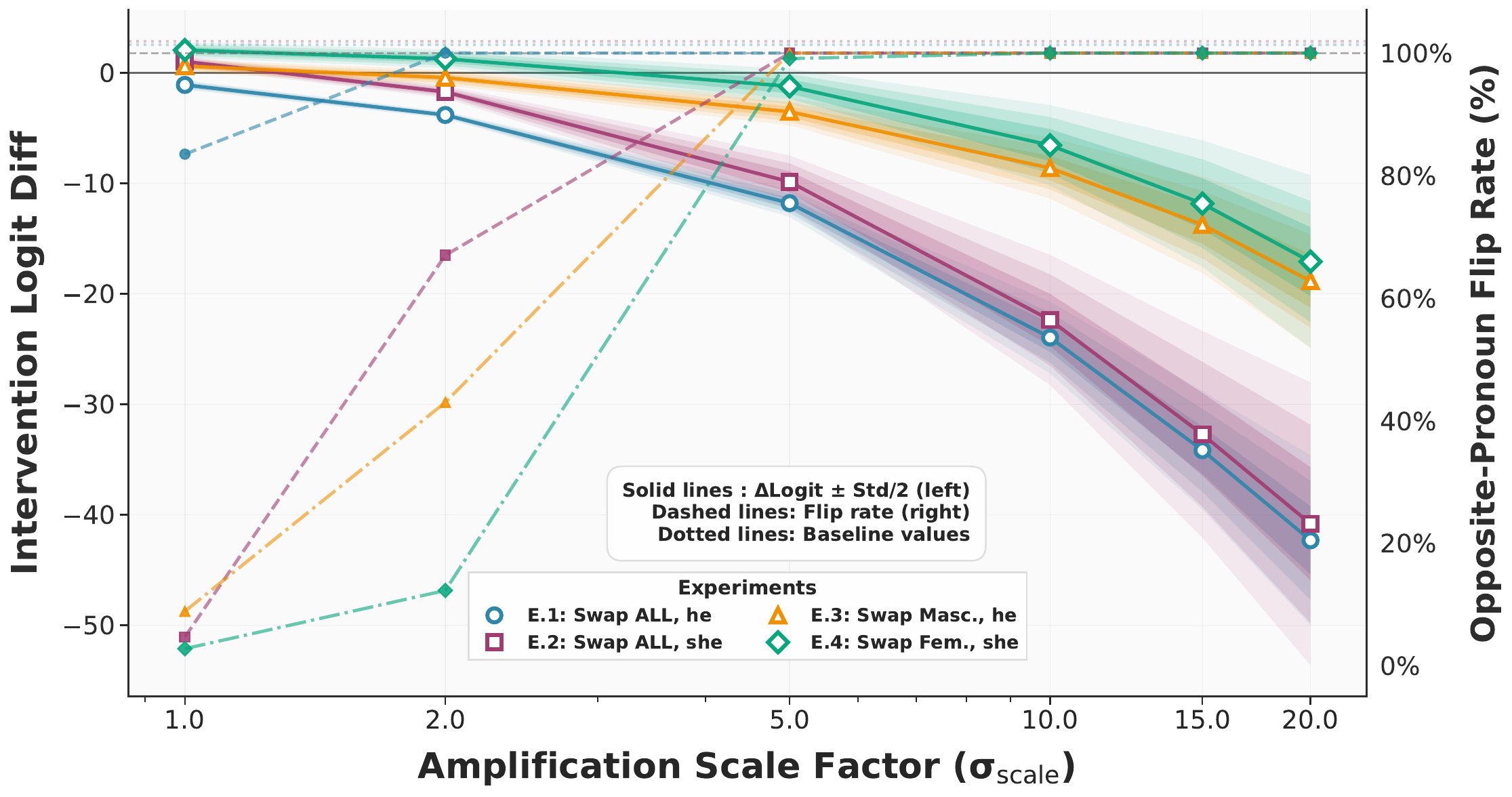}
    \caption{Causal interventions (scaling + swapping) show that singular directions control gender pronoun prediction. The plot displays logit differences (Correct Pronoun $-$ Opposite Pronoun) and flipping rates after intervention. Singular values $\sigma_i$ are scaled by an integer factor ($\sigma_{\text{scale}}$) in $\sigma_i (\tilde{a_i} - \nu^\top u_i) v_i^\top$, leading to near-complete prediction reversal at higher scales. This provides causal evidence that these directions are key computational units underlying gender pronoun resolution.}
\label{fig:scale-intervention-plot}
\end{figure}

\paragraph{Findings.}
The resulting interventions, as highlighted in Table \ref{tab:intervention-flip}, Table \ref{tab:range-ablation}, and Figure \ref{fig:scale-intervention-plot}, yield a clear causal signal. 
In Table \ref{tab:intervention-flip}, the Flip$\to$she\% is defined as the (\# predictions flipped from \texttt{`\underline{ }he'} $\to$ \texttt{`\underline{ }she'}) / (\# baseline \texttt{`\underline{ }he'} predictions) $\times 100$, and Flip$\to$he\%  is defined analogously using the baseline \texttt{`\underline{ }she'} predictions, i.e., (\# predictions flipped from \texttt{`\underline{ }she'} $\to$ \texttt{`\underline{ }he'}) / (\# baseline \texttt{`\underline{ }she'} predictions) $\times 100$. Both of these metrics help capture the true flip rate of recoverable predictions (excluding cases where the model initially predicted an "other" token). 
Across amplification scales shown in both Table \ref{tab:intervention-flip} and Figure \ref{fig:scale-intervention-plot}, we find that swapping all gender-associated singular directions with their empirically opposite-gender means, and scaling the corresponding singular values by a positive amplification factor, reliably reverses the model’s pronoun-logit polarity. For masculine prompts, the mean logit difference shifts from approximately \(+2.5\) (favoring \texttt{`\underline{ }he'}) to around \(-42\), and for feminine prompts from \(+2.8\) to roughly  \(-41\). Moreover, as also reflected in the high flip rates in Table \ref{tab:intervention-flip}, the model almost entirely reverses its prediction given the same baseline context, i.e., prompts containing masculine cues (where the correct baseline prediction is \texttt{`\underline{ }he'}) yield \texttt{`\underline{ }she’} after the mean-swap plus amplification intervention, and symmetrically for feminine-context prompts.
The scalar-level ablations in Table \ref{tab:range-ablation} further confirm that both full and partial swaps induce substantial, systematic shifts in logit differences, demonstrating that these singular directions contribute additively and consistently to gender-pronoun resolution.
All these results provide causal evidence that the identified OV singular directions form \textit{modular, low-rank mechanisms} that directly control gender pronoun prediction, demonstrating that interpretable subfunctions are not merely correlational artifacts but causal building blocks of transformer computation, which leads to a new perspective on looking at/understanding predictions made by the model. 

We believe that as we move forward, this direction/perspective of decomposing/understanding model decisions will make them more interpretable, and a wider range of transparent conceptualization can be explored/understood in future works.

% --- DOCUMENT START ---
% \begin{document}

\begin{table*}[htbp]
\centering
\footnotesize % Use a smaller font for the large table
\renewcommand{\arraystretch}{1.1} % Add a bit of vertical spacing
\setlength{\tabcolsep}{4pt} % Reset column spacing (3pt was a bit tight)
\caption{Empirical results of \textbf{intervention(scaling + swapping)}. Baseline $\Delta$Logit shows original logit difference between correct and opposite pronoun prediction based on the prompt context. Intervention $\Delta$Logit shows the same after replacing activations of  $\nu^{\top}u$ with respective opposite gender means, and amplifying singular value by multiplying with $\mathbf{\sigma_{scale}}$.  
Flip rates show the baseline pronoun predictions that were switched to the opposite pronoun by the intervention. All $\Delta$Logit(Logit Difference) values are shown as mean $\pm$ std.\\
% \textbf{Corrected Flip Percentage Notations:}
% The Flip$\to$she\% is the (\# predictions flipped from 'he' $\to$ 'she') / (\# baseline 'he' predictions) $\times$ 100. And Flip$\to$he\%  is the (\# predictions flipped from 'she' $\to$ 'he') / (\# baseline 'she' predictions) $\times$ 100.  \\
% \textit{This shows the TRUE flip rate of recoverable predictions (not including 'other' token predictions).}
}
\label{tab:intervention-flip}
% This spec uses ONE 'L' column and 6 fixed-width columns.
% The 'L' column will expand.
% The column definition is: L c c r r r r (7 columns total)
\begin{tabularx}{\textwidth}{@{} L c c r r r r @{}}
\toprule
% The header row must have 7 items, separated by 6 '&'
% NO '&' at the beginning of the line.
\textbf{Experiment} & \textbf{\makecell[c]{$\mathbf{\sigma_{scale}}$}} & \textbf{\makecell[c]{Prompt Type}} &
\textbf{\makecell[r]{Baseline \\ $\Delta$Logit}} & \textbf{\makecell[r]{Interv. \\ $\Delta$Logit}} &
\textbf{\makecell[r]{Flip$\to$ \\ she\%}} & \textbf{\makecell[r]{Flip$\to$ \\ he\%}} \\
\midrule
% All data rows must also have 7 items, separated by 6 '&'
% NO '&' at the beginning of the line.
E.1: Swap ALL & 1.0 & "he" & $+2.53 \pm 1.48$ & $-1.10 \pm 0.83$ & 83.5\% & 0.0\% \\
E.1: Swap ALL & 2.0 & "he" & $+2.53 \pm 1.48$ & $-3.82 \pm 0.63$ & 100.0\% & 0.0\% \\
E.1: Swap ALL & 5.0 & "he" & $+2.53 \pm 1.48$ & $-11.80 \pm 2.50$ & 100.0\% & 0.0\% \\
E.1: Swap ALL & 10.0 & "he" & $+2.53 \pm 1.48$ & $-23.96 \pm 5.52$ & 100.0\% & 0.0\% \\
E.1: Swap ALL & 15.0 & "he" & $+2.53 \pm 1.48$ & $-34.16 \pm 7.50$ & 100.0\% & 0.0\% \\
E.1: Swap ALL & 20.0 & "he" & $+2.53 \pm 1.48$ & $-42.31 \pm 8.57$ & 100.0\% & 0.0\% \\
\addlinespace[3pt]
E.2: Swap ALL & 1.0 & "she" & $+2.84 \pm 2.12$ & $+1.01 \pm 0.96$ & 0.0\% & 4.7\% \\
E.2: Swap ALL & 2.0 & "she" & $+2.84 \pm 2.12$ & $-1.73 \pm 0.61$ & 0.0\% & 67.0\% \\
E.2: Swap ALL & 5.0 & "she" & $+2.84 \pm 2.12$ & $-9.87 \pm 4.08$ & 33.3\% & 100.0\% \\
E.2: Swap ALL & 10.0 & "she" & $+2.84 \pm 2.12$ & $-22.37 \pm 9.31$ & 33.3\% & 100.0\% \\
E.2: Swap ALL & 15.0 & "she" & $+2.84 \pm 2.12$ & $-32.72 \pm 13.04$ & 33.3\% & 100.0\% \\
E.2: Swap ALL & 20.0 & "she" & $+2.84 \pm 2.12$ & $-40.82 \pm 15.44$ & 33.3\% & 100.0\% \\
\addlinespace[3pt]
E.3: Swap Masc. & 1.0 & "he" & $+2.53 \pm 1.48$ & $+0.60 \pm 1.22$ & 8.9\% & 0.0\% \\
E.3: Swap Masc. & 2.0 & "he" & $+2.53 \pm 1.48$ & $-0.44 \pm 0.97$ & 43.0\% & 0.0\% \\
E.3: Swap Masc. & 5.0 & "he" & $+2.53 \pm 1.48$ & $-3.53 \pm 0.63$ & 100.0\% & 0.0\% \\
E.3: Swap Masc. & 10.0 & "he" & $+2.53 \pm 1.48$ & $-8.60 \pm 1.69$ & 100.0\% & 0.0\% \\
E.3: Swap Masc. & 15.0 & "he" & $+2.53 \pm 1.48$ & $-13.50 \pm 2.98$ & 100.0\% & 0.0\% \\
E.3: Swap Masc. & 20.0 & "he" & $+2.53 \pm 1.48$ & $-18.15 \pm 4.15$ & 100.0\% & 0.0\% \\
\addlinespace[3pt]
E.4: Swap Fem. & 1.0 & "she" & $+2.84 \pm 2.12$ & $+2.05 \pm 1.39$ & 0.0\% & 2.8\% \\
E.4: Swap Fem. & 2.0 & "she" & $+2.84 \pm 2.12$ & $+0.35 \pm 0.73$ & 0.0\% & 12.3\% \\
E.4: Swap Fem. & 5.0 & "she" & $+2.84 \pm 2.12$ & $-4.75 \pm 1.92$ & 16.7\% & 99.1\% \\
E.4: Swap Fem. & 10.0 & "she" & $+2.84 \pm 2.12$ & $-13.04 \pm 5.59$ & 50.0\% & 100.0\% \\
E.4: Swap Fem. & 15.0 & "she" & $+2.84 \pm 2.12$ & $-20.73 \pm 8.82$ & 50.0\% & 100.0\% \\
E.4: Swap Fem. & 20.0 & "she" & $+2.84 \pm 2.12$ & $-27.59 \pm 11.44$ & 50.0\% & 100.0\% \\
\bottomrule
\end{tabularx}
\end{table*}

\begin{table*}[t]
\centering
\caption{
Scalar-based ablation of gender-related OV singular directions.
Each intervention replaces the scalar activation of each gender direction with its
empirical opposite-gender mean and amplifying the singular value($\times20$ in below table). $n$ is the number of data points
% All interventions cause large, statistically significant reversals in logit differences
% ($|\Delta|\!\approx\!40$ for full swaps).
% Masculine directions act consistently across contexts (low variance, $\sigma\!\approx\!3.2$),
% while feminine directions are more variable ($\sigma\!\approx\!9.9$),
% indicating stronger context dependence and interaction effects.
}
\footnotesize
\renewcommand{\arraystretch}{1.0}
\setlength{\tabcolsep}{6pt}
\begin{tabularx}{\textwidth}{@{}l c c c c c@{}}
\toprule
\textbf{Experiment} & \textbf{Prompt Context} & $n$ &
Baseline $\Delta$Logit & Interv.$\Delta$Logit & $\Delta$($\Delta$Logit) \\
\midrule
E.1: Swap ALL dirs & ``he'' & 150 &
$+2.53\pm1.48$ & $-42.31\pm8.57$ & $-44.84$ \\[2pt]
E.2: Swap ALL dirs & ``she'' & 156 &
$+2.84\pm2.12$ & $-40.82\pm15.44$ & $-43.66$ \\[2pt]
E.3: Swap Masc.~only & ``he'' & 150 &
$+2.53\pm1.48$ & $-18.15\pm4.15$ & $-21.38$ \\[2pt]
E.4: Swap Fem.~only & ``she'' & 156 &
$+2.84\pm2.12$ & $-27.59\pm11.44$ & $-19.93$ \\
\bottomrule
\end{tabularx}
\label{tab:range-ablation}
\end{table*}

\begin{table}[t]
\centering
\caption{The table shows results for the analysis of $S_7$ attention scores($[1, x_i]\sigma_7u_7v_7^\top[1,x_j]^\top$) for head 9.6, highlighting the entity-action distinction role in the IOI task.}
\begin{tabular}{lccc}
\toprule
\textbf{Category} & \textbf{Value (Mean $\pm$ SD)} & \textbf{Example tokens} & \textbf{Example values} \\
\midrule
Entities & $+3.52 \pm 1.42$ & "Jerry", "Kevin", "Susan" & $+2.87$, $+2.29$, $+2.92$ \\
Actions & $-4.44 \pm 0.68$ & "went", "gave", "decided" & $-4.33$, $-4.17$, $-5.93$ \\
\bottomrule 
\label{tab:entity_action_seperation}
\vspace{-3mm}
\end{tabular}
\end{table}

\begin{table}[t]
\centering
\caption{The Table shows results for analysis of $S_{28}$ activations, denoting the role of entity salience in the IOI task.}
\begin{tabular}{lcc}
\toprule
\textbf{Token type} & \textbf{Example tokens} & \textbf{SVD\_28 values} \\
\midrule
Named entities (first mention) & "Jerry", "Susan", "Kevin" & $5.68$, $4.05$, $5.22$ \\
Named entities (second mention) & "Mary", "Kevin\_2", "Marilyn\_2" & $3.69$, $2.07$, $2.07$ \\
Function words & "the", "to", "of" & $0.50$, $1.93$, $0.15$ \\
Discourse connectives & "and" & $2.92$-$3.85$ \\
\bottomrule 
\label{tab:entity_detection}
\vspace{-7mm}
\end{tabular}
\end{table}

\subsection{Extended Analysis: IOI Task}
\label{app:ioi-extended}

\begin{figure*}  % figure* spans multiple columns, [p] for page float
    \centering
    \makebox[\textwidth][c]{  % centers and allows content wider than \textwidth
        \includegraphics[width=0.9\linewidth]{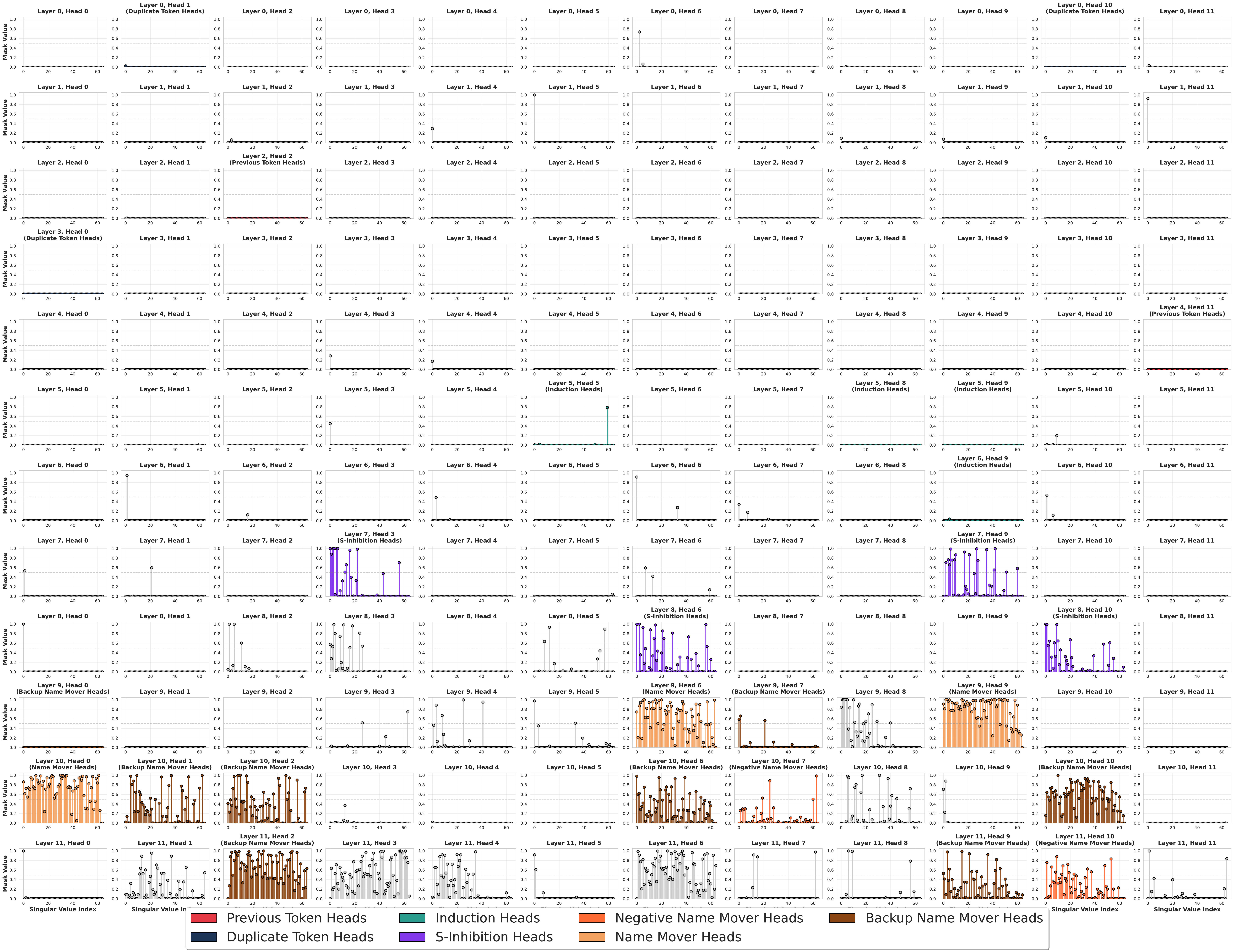}
    }
    \caption{Learned singular value masks for OV ($\mathbf{W_{aug}^{OV}}$) matrices across all attention heads in the model. The masks show heads with high activation across multiple singular dimensions correspond to circuit components previously identified by \citet{wang2022interpretabilitywildcircuitindirect} for the IOI task.}
    \label{fig:ov_singular_masks}
\end{figure*}

\begin{figure*}  % figure* spans multiple columns, [p] for page float
    \centering
    \makebox[\textwidth][c]{  % centers and allows content wider than \textwidth
        \includegraphics[width=0.4\linewidth]{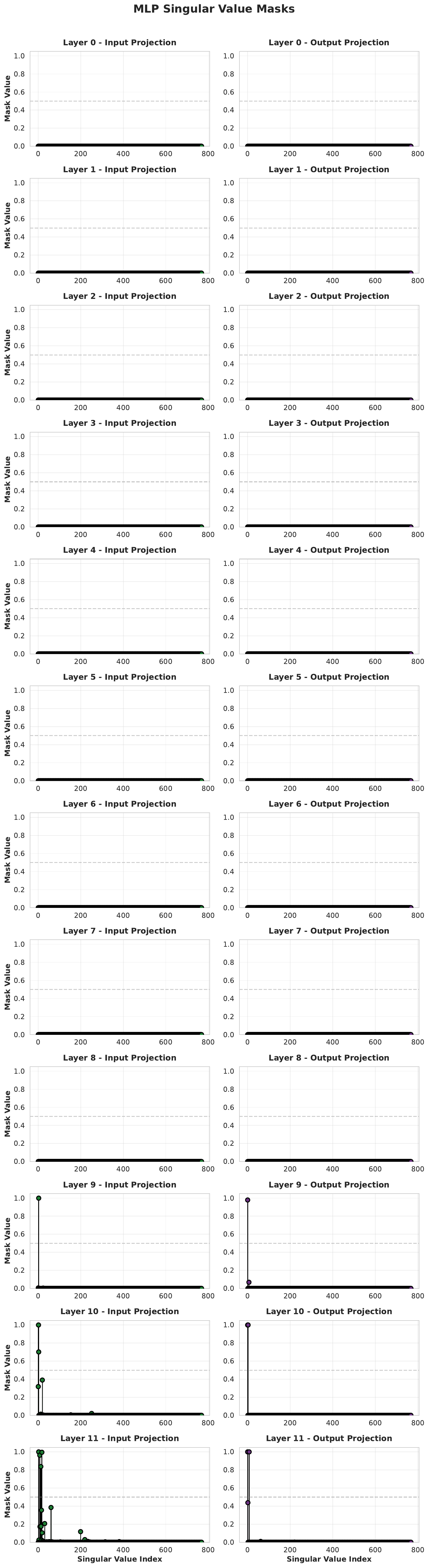}
    }
    \caption{Learned singular value masks for MLP ($\mathbf{W_{aug}^{(in)}}\text{(left) and }\mathbf{W_{aug}^{(out)}}$(right)) matrices across all layers in the model for the IOI task. }
    \label{fig:mlp_singular_masks}
\end{figure*}

\begin{figure}[t]
    \centering
    \includegraphics[width=0.89\linewidth]{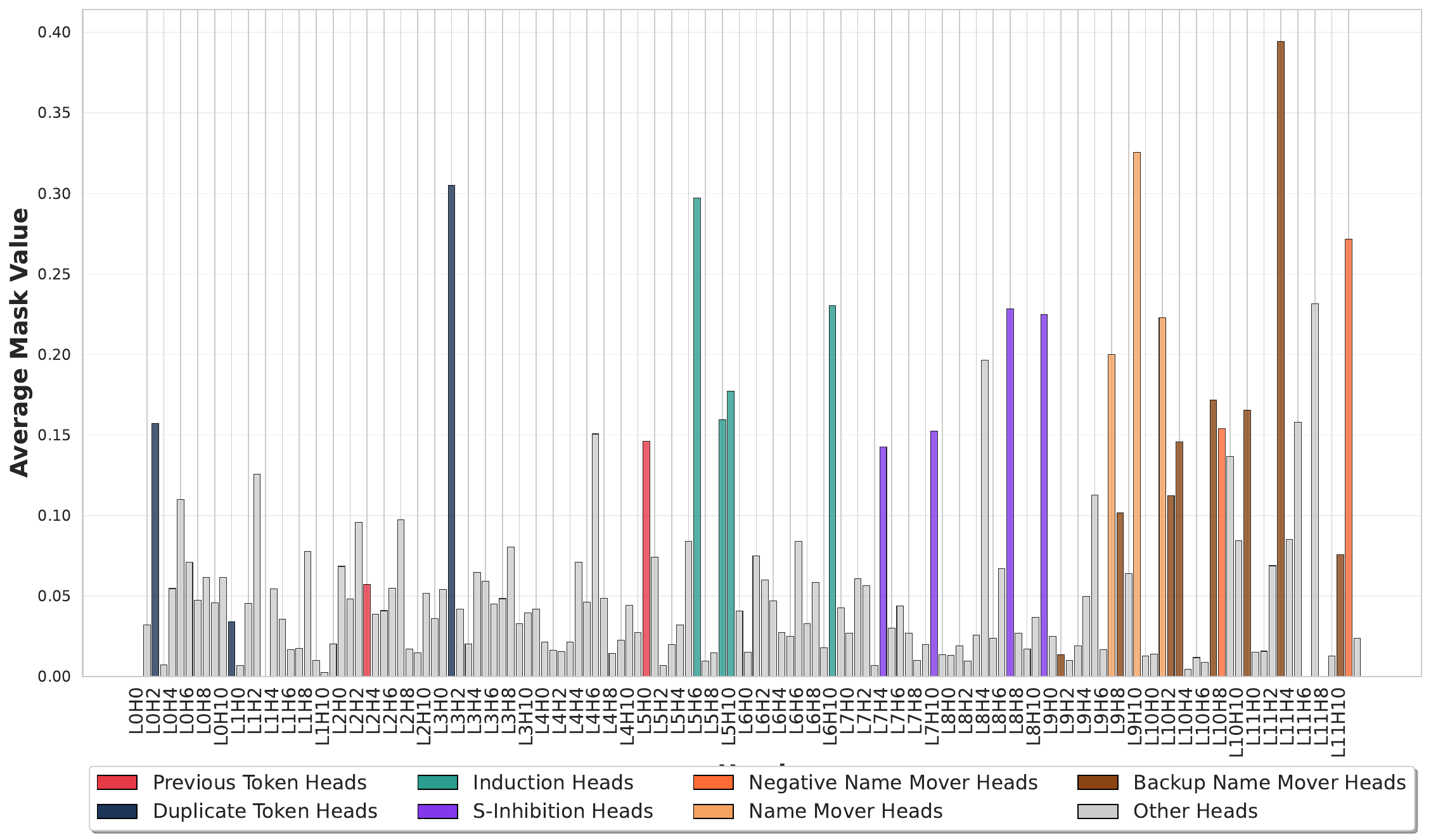}
    % % \vspace{-3mm}
    \caption{The figure shows the average mask values for $W_{aug}^{QK}$ across attention heads, categorized by functional type. Heads identified in circuits by \citet{wang2022interpretabilitywildcircuitindirect}; particularly Name Mover, Backup Name Mover, and Negative Name Mover heads, consistently exhibit higher average mask values than other head types. This suggests a correlation between circuit membership and mask activation strength, providing quantitative validation of previously identified functional circuits.}
    \label{fig:attention-mask-distribution}
    % % \vspace{-2mm}
\end{figure}

\begin{figure}
    \centering
    \includegraphics[width=\linewidth]{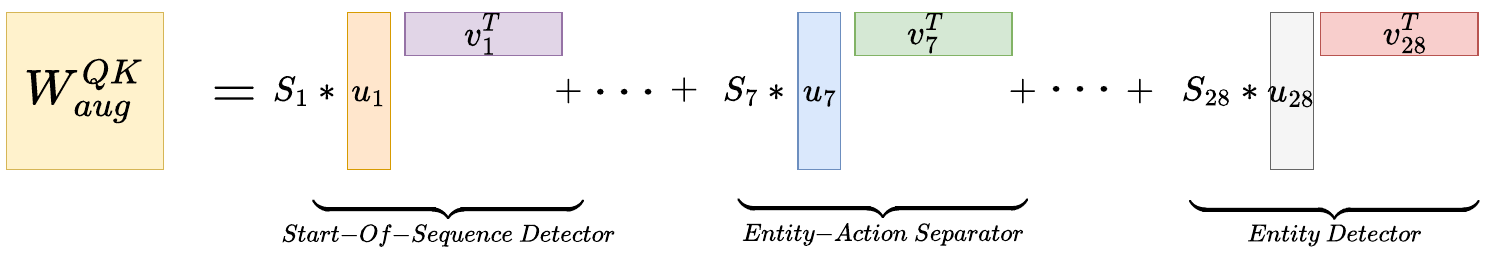}
    % \vspace{-3mm}
    \caption{Analysis of head 9.6, previously identified by \citet{wang2022interpretabilitywildcircuitindirect} as a ``Name Mover head'' that attends to previous names in a sentence and copies them. Our SVD analysis reveals multiple distinct functionalities within the $QK$ interaction, each serving specific roles consistently across the dataset. This decomposition provides a more nuanced understanding of the head's behavior.}
    \label{fig:function-svd-9.6}
    % \vspace{-4mm}
\end{figure}

\paragraph{Additional Mask Visualizations.}
Figures \ref{fig:ov_singular_masks}, \ref{fig:mlp_singular_masks}, and \ref{fig:attention-mask-distribution} provide detailed visualizations of the learned singular value masks for OV, MLP, and QK matrices, respectively. These figures complement the results discussed in the main text by showing head- and layer-level activation patterns across the model. In particular, Figure \ref{fig:ov_singular_masks} highlights OV heads with high activation across multiple singular directions, Figure \ref{fig:mlp_singular_masks} shows layer-wise MLP mask patterns, and Figure \ref{fig:attention-mask-distribution} quantitatively correlates QK mask values with functional head types identified in prior work \citep{wang2022interpretabilitywildcircuitindirect}. These visualizations help provide further evidence that the learned masks capture meaningful functional subcomponents of transformer computation. Figure \ref{fig:function-svd-9.6} shows a conceptualization of the SVD-based analysis of Head 9.6, previously identified by \citet{wang2022interpretabilitywildcircuitindirect} as a “Name Mover” head. The figure illustrates the combination of multiple distinct functionalities within the $QK$ interaction, including sequence initialization (S1), semantic discrimination (S7), and entity salience (S28), leading to a specific utility in the IOI task.

\begin{table}[t]
\centering
\caption{Learned mask values and associated functional roles for selected singular directions in \texttt{Layer 9, Head 6}. Higher mask values indicate a stronger contribution of that singular direction to the corresponding functional role, illustrating how specific QK directions encode distinct computational subfunctions.}
\begin{tabular}{ccc}
\toprule
\textbf{Singular Direction} & \textbf{Functional Role} & \textbf{Learned Mask Value} \\ \midrule
$S_{1}$  & Sequence Initialization Detection & 0.53 \\ 
$S_{7}$  & Semantic Separation of Entities and Actions  & 0.64  \\ 
$S_{28}$ & Entity Salience and Detection & 0.97 \\ \bottomrule
\end{tabular}
\label{tab:functional_role}
\end{table}

To complement the in-depth analysis of attention head 9.6 in the main paper, we extend our investigation to two additional heads, 9.9 and 10.0, both previously implicated in the Indirect Object Identification (IOI) task by \citet{wang2022interpretabilitywildcircuitindirect}. Our aim is to determine whether their internal structure, when decomposed through our singular vector masking framework, reveals similarly modular and interpretable subfunctions. The results provide compelling support for the view that heads encode multiple, independent functional primitives, each realized along distinct low-rank directions.

\paragraph{Head 9.9: Entity Prioritization and Inhibition Dynamics}
Attention head 9.9 has been previously labeled a ``Name Mover'' head due to its role in copying names across syntactic spans. However, our singular vector analysis reveals a richer and more nuanced structure. The first and seventh singular directions, $S_1$ and $S_7$, consistently exhibit the characteristic pattern of start-of-sequence detectors. These directions assign disproportionately high attention to the first token of each prompt, while suppressing all subsequent tokens. This behavior mirrors similar mechanisms found in other heads and suggests a general strategy for anchoring temporal computations.

More interestingly, the $51^{\text{st}}$ direction, $S_{51}$, exhibits a selective affinity for named entities, especially those appearing early in a sequence. It systematically favors tokens corresponding to person names while suppressing their repeated mentions. This results in a subtle form of repetition avoidance that biases the model toward novel or contextually salient entities, behavior consistent with entity tracking and attention modulation observed in prior mechanistic studies.

Crucially, $S_{16}$ implements a strong inhibitory signal targeting second mentions of entities. On average, the unnormalized attention score assigned to repeated entities is $2.31$ points lower than that assigned to their initial mention, a pattern aligned with the role of S-Inhibition heads described by \citet{wang2022interpretabilitywildcircuitindirect}, which helps suppress the attention. This directional suppression creates a structural preference against re-attending to previously mentioned entities, thereby enhancing the likelihood of selecting an appropriate indirect object. Collectively, these observations demonstrate that head 9.9 multiplexes multiple functions, entity recognition, novelty bias, and inhibition, across orthogonal subspaces.

\paragraph{Head 10.0: Semantic Role Differentiation and IOI Bias Encoding}

Head 10.0 presents another compelling case of functional modularity. As with head 9.9, the directions $S_1$ and $S_5$ operate as strong start-of-sequence indicators, highlighting the recurrence of this structural primitive across layers. However, the sixth singular direction, $S_6$, shows an especially focused pattern: it consistently assigns high attention scores to named entities and location nouns, while strongly de-emphasizing function words and verbs. For instance, tokens such as “Melissa,” “Kelly,” and “zoo” receive the highest activations in our sampled prompts, whereas verbs like “gave” and conjunctions like “and” are suppressed. This direction functions as a precise semantic filter, elevating entities and key nouns that anchor the core referential structure of the IOI task.

In contrast, the third direction, $S_3$, displays a consistent aversion to function words. Words such as “a,” “the,” and “and” receive markedly negative attention scores, suggesting a broader mechanism of syntactic sparsification. Meanwhile, the seventh direction, $S_7$, echoes the behavior observed in head 9.6, acting as a semantic separator between entities and verbs. These orthogonal semantic dimensions help disentangle the “who” from the “what,” facilitating downstream resolution of coreference.

Finally, the twentieth singular direction, $S_{20}$, exhibits a statistical preference for indirect objects. In approximately 54–59\%  of cases, this direction assigns higher attention to the token corresponding to the indirect object, closely mirroring the empirical success rates of the model on the IOI task. This suggests that $S_{20}$ encodes a partial bias that systematically favors the correct grammatical resolution in ambiguous syntactic constructions.

\paragraph{Overall Summary:}

The analyses of heads 9.9 and 10.0 reinforce our core hypothesis, i.e., transformer components, rather than operating as indivisible functional units, exhibit finely grained internal specialization along interpretable low-rank directions. Each direction performs a targeted role, whether syntactic anchoring, semantic discrimination, or statistical preference modulation, that contributes to the overall behavior of the head. Our results not only validate prior circuit-level findings but also refine them by isolating the precise mechanisms responsible for observed model behavior. This decomposition presents a powerful perspective for interpretability, enabling a move from coarse component-level attributions to direction-level mechanistic understanding.

\subsection{Analysis for \textit{Greater Than} task}
\label{app:gt-analysis}

To evaluate the generality of our method beyond the IOI task, we extend our analysis to the \textit{Greater Than} benchmark introduced by \citet{hanna2023does}, which investigates a model’s capacity for numerical comparison. Each input prompt presents two years within a templated sentence, e.g., \textit{``The treaty lasted from the year 1314 to the year 13''}, and the model must complete the final token(s) such that the resulting year is strictly greater than the first. Crucially, the completion must yield a valid multi-token year (e.g., \textbf{28} completing ``1328’’), with careful curation to avoid boundary conditions and single-token years that could confound interpretability.

We analyze this task by focusing on attention heads previously identified as critical to the model’s numerical comparison behavior, specifically heads 6.9, 9.1, and 5.5 in GPT-2. Using our method, we dissect the $W_{\text{aug}}^{QK}$ matrices of these heads into their dominant singular directions and interpret their respective contributions.

\begin{table}[t]
\centering
\caption{SVD Component Analysis of $\mathbf{W_{\text{aug}}^{QK}}$ Attention for the \textit{Greater Than} Task ,  Focus on Head 9.1. Multiple low-rank components exhibit high attention to the target year token ($YY$), which is crucial for accurate prediction. The “Highest Attention \%” indicates how often $YY$ received the highest attention score.}
\label{tab:svd_9.1_gt_stats}
\begin{tabular}{cccc}
\toprule
% \hline
\textbf{SVD Component} & \textbf{Avg. Attention ($\pm$ Std)} & \textbf{Highest Attention (\%)}& \textbf{Mask Value} \\
% \hline
\midrule
$S_{31}$ & $7.09 \pm 1.68$  & 100 & 1.00\\
% \hline
$S_3$   & $4.79 \pm 1.08$  & 99.8 & 1.00 \\
% \hline
$S_{26}$ & $1.86 \pm 0.87$  & 90.9 & $3.29\times 10^{-5}$\\
% \hline
$S_{59}$ & $2.94 \pm 1.86$  & 81.8 & 1.00 \\
% \hline
$S_{37}$ & $4.71 \pm 2.18$ & 81.5 & 1.00 \\
% \hline
$S_{32}$ & $3.90 \pm 1.65$  & 72.7 & 1.00\\
% \hline
% $S_{53}$ & $5.19 \pm 2.62$  & 45.5 \\
% \hline
\bottomrule
\end{tabular}
\end{table}

\begin{figure}
    \centering
    \includegraphics[width=\linewidth]{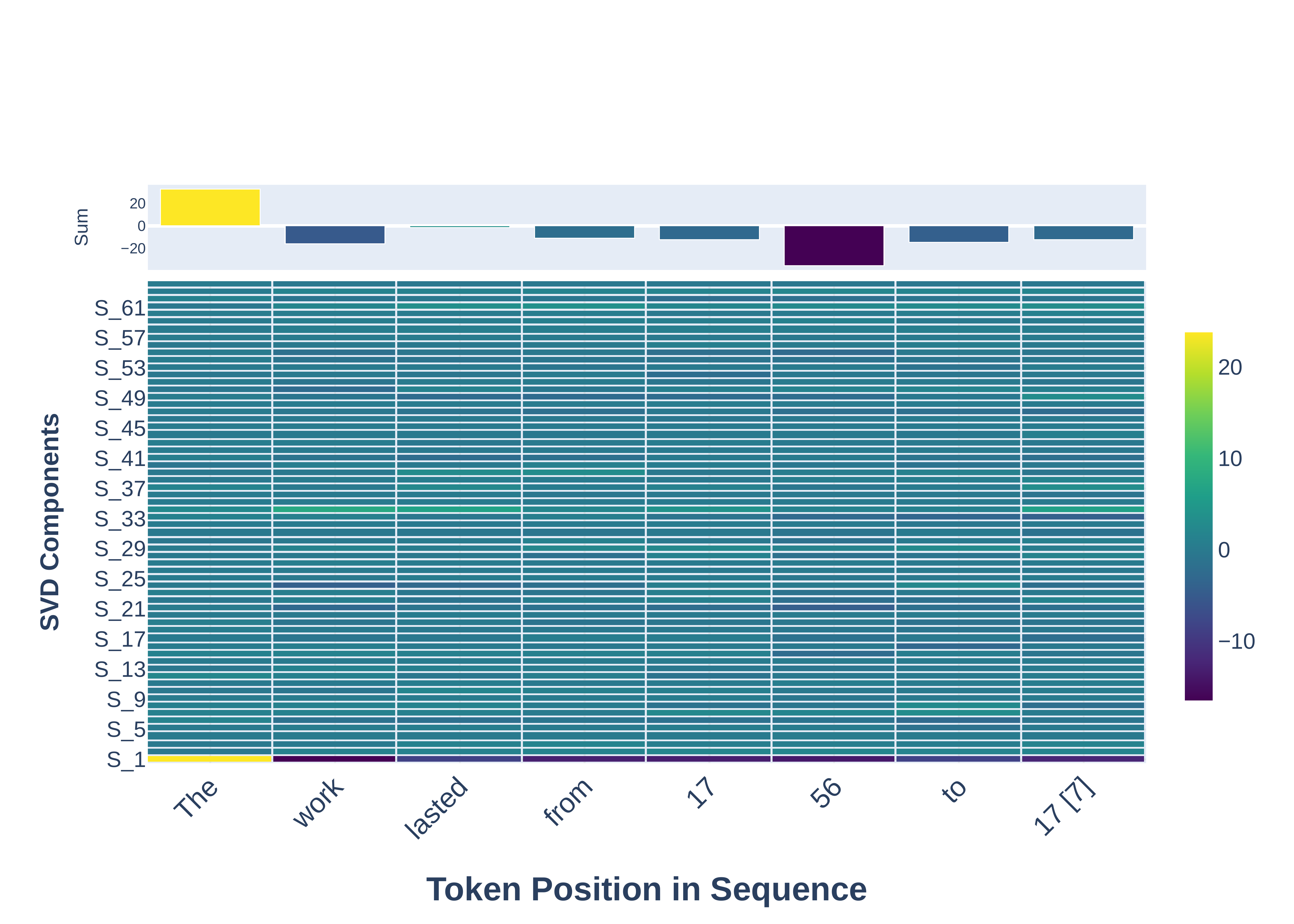}
    \caption{The above figure shows the attention score of head 9.1 for the \textit{Greater Than} task. $S_1$ attends highly to the first token, and others, such as $S_3$ and $S_{31}$, predominantly focus on the end-of-year token.}
    \label{fig:gt-head-heatmap}
\end{figure}

\paragraph{Head 9.1: Distributed Temporal Anchoring}

Attention head 9.1 (Figure \ref{fig:gt-head-heatmap}) reveals a distributed mechanism for endpoint detection. The first singular direction ($S_1$) consistently assigns high positive attention to the start-of-sequence token, while giving negative scores to subsequent ones. This behavior functions as a positional prior, commonly reused across tasks.

More importantly, multiple directions contribute to highlighting the terminal year token ($YY$). For instance, $S_{31}$ and $S_3$ achieve near-perfect precision in identifying $YY$ as the attention focus, doing so in 100\% and 99.8\% of prompts, respectively (see Table~\ref{tab:svd_9.1_gt_stats}). Other components such as $S_{32}$ and $S_{59}$ also consistently amplify $YY$, although with lower magnitude or frequency. This suggests that head 9.1 does not rely on a single dominant axis, but instead employs a compositional mechanism where multiple orthogonal components redundantly encode attention to temporal anchors.

\begin{table}[t]
\centering
\caption{SVD Component Analysis of $\mathbf{W_{\text{aug}}^{QK}}$ Attention for the \textit{Greater Than} Task ,  Focus on Head 6.9. Multiple low-rank components exhibit high attention to the target year token ($YY$), which is crucial for accurate prediction. The “Highest Attention \%” indicates how often $YY$ received the highest attention score.
% \AJ{add learned mask value to table}
}
\label{tab:svd_6.9_gt_stats}
\begin{tabular}{cccc}
% \hline
\toprule
\textbf{SVD Component} & \textbf{Avg. Attention $\pm$ Std} & \textbf{Highest Attention (\%)} & \textbf{Mask Value}\\
% \hline
\midrule
$S_2$ & $10.747 \pm 2.858$ & $100.0$ & 1.00\\
% \hline
$S_7$ & $4.324 \pm 2.482$ & $83.3$ & $6.81\times10^{-6}$\\
% \hline
$S_{18}$ & $4.680 \pm 2.327$ & $83.3$ & 0.15\\
% \hline
$S_{20}$ & $4.897 \pm 4.967$ & $66.7$ & 0.99\\
% \hline
$S_{29}$ & $3.249 \pm 3.587$ & $50.0$ & $1.14\times10^{-5}$\\
% \hline
% $S_{49}$ & $4.810 \pm 2.191$ & $16.7$ \\
% \hline
\bottomrule
\end{tabular}
\end{table}

\paragraph{Head 6.9: Sharply Localized Numerical Discrimination}
In contrast, head 6.9 demonstrates highly concentrated behavior. The second singular direction, $S_2$, exhibits extremely sharp selectivity, consistently assigning the highest attention to the final year token with 100\% accuracy ($\mu = 10.747 \pm 2.858$) (see Table \ref{tab:svd_6.9_gt_stats} for reference). This direction appears to isolate the second numeric entity in the input, crucial for the Greater Than judgment. Supporting directions such as $S_7$, $S_{18}$, and $S_{20}$ reinforce this emphasis with selection rates above 66\%, albeit at lower attention magnitudes. This pattern reflects a localized, narrowly targeted strategy for operand comparison.

\paragraph{Head 5.5: Redundant High-Magnitude in Reinforcing the Endpoint}

Head 5.5 features a richer distribution of strong attention-inducing components. The leading singular direction ($S_1$) again exhibits the start-of-sequence pattern, while several others like $S_{47}$, $S_{28}$, $S_{52}$, robustly identify the $YY$ token. Notably, $S_{15}$ stands out with the highest attention magnitude across all heads ($19.14 \pm 7.15$), suggesting an exceptionally focused mechanism for endpoint amplification.

While some of these components exhibit redundancy in their selection behavior (e.g., $S_{28}$ and $S_{52}$ both achieving 87.5\% attention on $YY$), they differ significantly in their mean scores and variance. This dispersion implies an ensemble encoding, where multiple vectors provide convergent evidence toward the correct numeric endpoint, increasing the model’s reliability across input variations.

\paragraph{Cross-Head Trends and Shared Functionalities}

Across all heads, we observe the emergence of two recurring functional patterns: (1) start-of-sequence structure detection via $S_1$, and (2) endpoint amplification via a sparse set of highly specialized directions. These shared functionalities suggest that certain subspaces, such as those capturing positional priors or token-type distinctions (e.g., numbers vs. nouns), may be reused across attention contexts. Moreover, the existence of multiple low-rank directions targeting the same token suggests that GPT-2 distributes the implementation of numerical reasoning across a sparse ensemble of interpretable basis vectors.

\paragraph{Overall Summary:}
This decomposition of the Greater Than task reveals a consistent, structured strategy whereby numerical comparison is encoded not in monolithic head-level behavior, but in sparse, orthogonal subspaces within each attention matrix. Each direction contributes a distinct yet complementary role, which includes establishing context, identifying operand positions, and assigning salience to temporally relevant entities. These findings provide compelling evidence that transformers perform symbolic reasoning via emergent low-dimensional structures, and that these structures are modular, reusable, and interpretable via singular decomposition.

\subsection{Universal Composite Functionalities Discovered}
\label{com-func-details}

\begin{figure}
    \centering
    \includegraphics[width=0.9\linewidth]{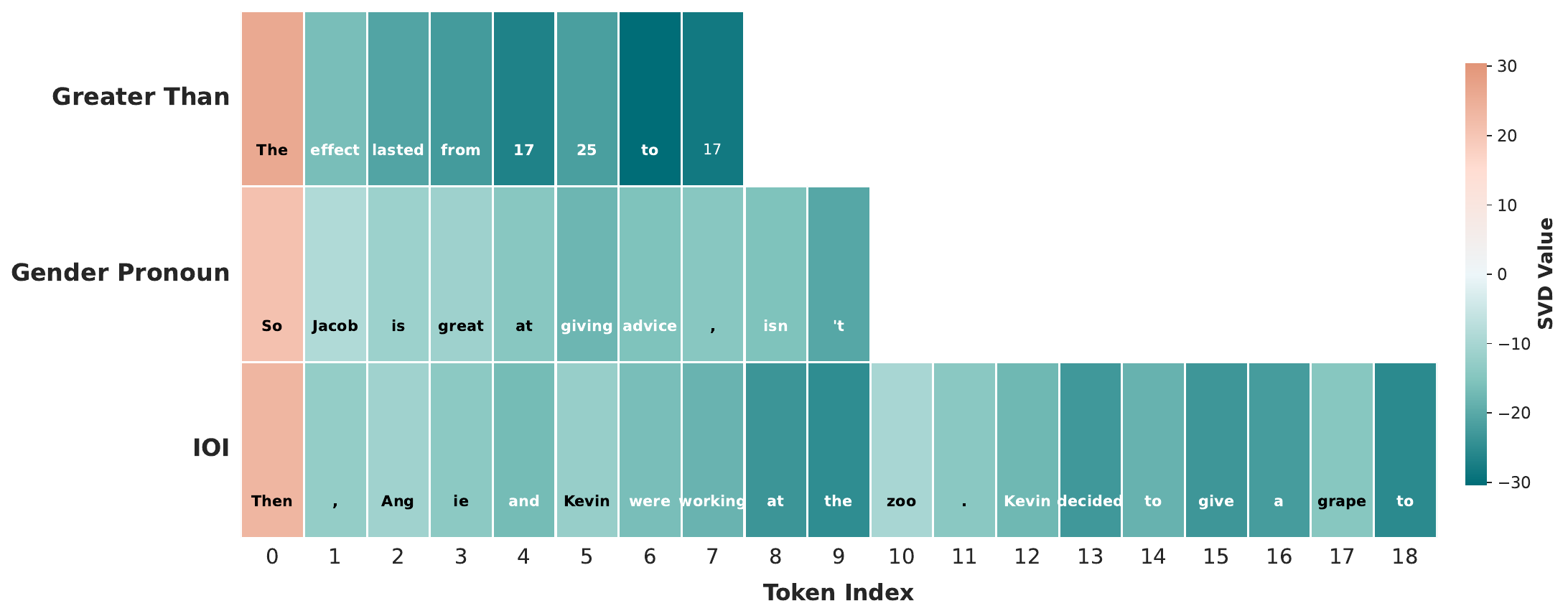}
   \caption{The figure above shows an instance of attention scores of the final token for component $S_1$ of head 9.6. It demonstrates that $S_1$
  consistently functions as a start-of-sequence detector across tasks, independent of context.}
    \label{fig:univeral-start-of-seq-detector}
\end{figure}

\begin{figure}
    \centering
    \includegraphics[width=\linewidth]{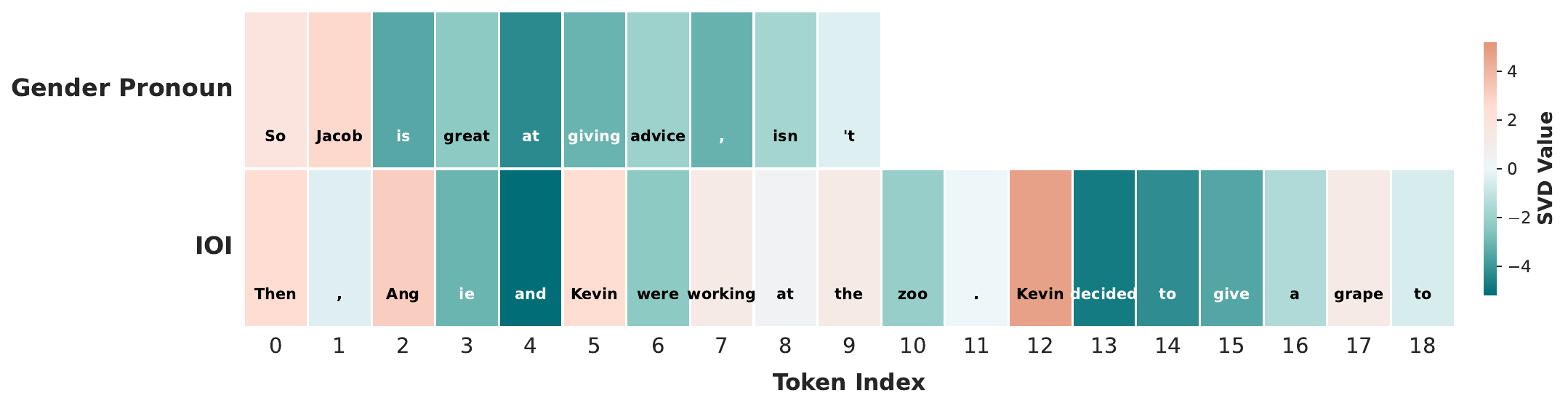}
    \caption{The figure above shows an instance of attention scores of the final token for component $S_7$ of head 9.6. It demonstrates that $S_7$
  consistently act as an Entity action separator by giving the highest attention score to Name and object entities, and the least attention score to actions.}
    \label{fig: universal-entity-action-seperator}
\end{figure}

Our analysis reveals that certain singular value components exhibit consistent, reusable functionalities across different attention heads and tasks, pointing to the presence of universal composite functionalities within GPT-2’s internal representations. For example, component $S_1$ in attention head 9.6 consistently acts as a \textit{start-of-sequence detector}, robustly assigning high attention scores to the initial token regardless of the task context, as illustrated in Figure~\ref{fig:univeral-start-of-seq-detector}. This persistent role suggests that some attention subspaces are dedicated to foundational structural signals critical for sequence processing. Similarly, component $S_7$ of the same head functions reliably as an \textit{entity-action separator}, selectively attending to name and object entities while suppressing attention to action or functional tokens, as shown in Figure~\ref{fig: universal-entity-action-seperator}. The consistency of these components across diverse contexts implies that transformer models implement a set of primitive, composable mechanisms that are repurposed modularly to support a variety of reasoning and linguistic tasks. Uncovering and characterizing such universal functionalities not only enriches our understanding of transformer interpretability but also paves the way for targeted interventions, modular editing, and transfer of learned behaviors across models and domains. These findings motivate further investigation into other universal components and their interplay, which may reveal a hierarchical structure of model reasoning primitives embedded in singular value decompositions of attention matrices.

\subsection{Sparsity Computation}
\label{app-sec:sparsity}

We define sparsity in terms of the number of singular directions that are effectively suppressed during the mask learning process.
Prior to optimization, all directions with negligible singular values are discarded, leading to a negligible
penalty of approximately $10^{-6}$ in KL divergence.
The remaining directions constitute the learnable subspace over which mask optimization is performed.

Since the model undergoes two kinds of compression, 
(i) Zeroing out directions with near-zero singular values, and
(ii) Pruning directions with high singular values through learned masks,  we report two complementary sparsity measures:
\paragraph{Relative Sparsity.}
    Let $n_{\text{active}}$ denote the number of singular directions having greater mask value then the $threshold(1\times10^{-2})$ after mask optimization, and
    $N_{\text{learnable}}$ the total number of directions subject to training.
    Then the relative sparsity is
    \[
        S_{\text{rel}} = 1 - \frac{n_{\text{active}}}{N_{\text{learnable}}}.
    \]
    This measures sparsity within the learnable subset only.

\paragraph{Full Sparsity.}
    Considering the complete model, let $N_{\text{total}}$ be the total number of singular directions available
    across all OV projections. The full sparsity is defined as
    \[
        S_{\text{full}} = 1 - \frac{n_{\text{active}}}{N_{\text{total}}}.
    \]
    This reflects overall model compression after both truncation and pruning.

% \subsection{Distribution}
% \label{app-sec:OOD}
\subsection{Discussion and Future Directions}
\label{app:future-directions}

Our proposed decomposition-based framework presents a scalable, model-agnostic approach for discovering interpretable subspaces within pretrained transformers. By applying singular value decomposition to attention weight matrices, we identify low-rank directions that robustly encode functional roles, such as temporal endpoint detection or entity repetition suppression. This enables fine-grained dissection of emergent behaviors, such as the "Name Mover" or "Greater Than" circuits, and reveals that these are implemented through sparse combinations of reusable subspaces rather than monolithic structures. Notably, we observe recurring components across tasks, such as start-of-sequence detectors and temporal amplifiers, suggesting that transformer models like GPT-2 leverage a compact set of primitive functions in compositionally rich ways. Our results align with and extend prior mechanistic interpretability research, complementing methods such as contextual decomposition \citep{hsu2025efficient}, attribution patching \citep{nanda2023attribution}, and circuit overlap metrics \citep{hanna2024have}, while providing a new axis of interpretability rooted in linear decomposability. The approach is computationally efficient, all analyses were performed on a single NVIDIA A40 GPU, and it reveals that many key behaviors are localized to a small number of interpretable singular directions. This raises intriguing possibilities for circuit editing, model compression, or steering via low-rank intervention. However, our results are not exhaustive; rather, they provide a first step toward uncovering modular internal mechanisms through decomposition. We emphasize that this work identifies functionalities that are interpretable under this perspective, but there is much more to uncover. Future directions include conducting targeted intervention experiments to validate causal contributions of discovered components, examining whether these mechanisms generalize to larger or more capable models (e.g., Phi, LLaMA), and exploring their activation in real-world settings. Additionally, the search for composite patterns, i.e., higher-order patterns composed of recurring singular directions, may yield deeper insight into how transformers orchestrate symbolic reasoning.

\end{document}